\documentclass[conference]{IEEEtran}

\newif\ifcameraready
\ifdefined\camerareadyoff
  \camerareadyfalse
\else
  \camerareadytrue
\fi

\usepackage{tikz}
\usetikzlibrary{calc}
\usepackage{verbatim}
\usepackage{makecell}

\usepackage{graphicx}
\usepackage{amsmath}
\usepackage{amssymb}
\usepackage{amsthm}
\usepackage{xcolor}
\usepackage{xspace}
\usepackage{bm}
\usepackage{booktabs}
\usepackage{multirow}
\usepackage{siunitx}
\usepackage{subcaption}
\usepackage{enumitem}
\usepackage[linesnumbered, ruled, vlined]{algorithm2e}
\usepackage{soul}
\usepackage[normalem]{ulem}

\makeatletter
\DeclareRobustCommand\onedot{\futurelet\@let@token\@onedot}
\def\@onedot{\ifx\@let@token.\else.\null\fi\xspace}

\def\etal{\emph{et al}\onedot}
\makeatother

\definecolor{bycolor}{rgb}{0.85,0.42,0.00}% orange
\definecolor{qlcolor}{rgb}{0.00,0.45,0.65}% cyan

\definecolor{revcolor}{rgb}{0.05,0.20,0.70}% blue: revised text, diff off
\definecolor{revdelcolor}{rgb}{0.62,0.05,0.05}% dark red: removed text
\definecolor{revaddcolor}{rgb}{0.00,0.42,0.16}% dark green: added text
\newif\ifrevisionshowdiff
\ifdefined\revisiondiffon\revisionshowdifftrue\else\revisionshowdifffalse\fi
\long\def\revisionnotext{}
\DeclareRobustCommand{\revisionmarkold}[1]{%
  \begingroup
    \long\def\revisionoldtext{#1}%
    \ifx\revisionoldtext\revisionnotext\else{\color{revdelcolor}#1}\space\fi
  \endgroup
}
\DeclareRobustCommand{\revision}[2]{%
  \ifcameraready
    #2%
  \else
    \ifrevisionshowdiff
      \revisionmarkold{#1}{\color{revaddcolor}#2}%
    \else
      {\color{revcolor}#2}%
    \fi
  \fi
}

\usepackage{bbm}

{\begin{list}               %
    {$\bullet$ \hfill}{
        \setlength{\leftmargin}{\parindent}
        \setlength{\parsep}{0.04\baselineskip}
        \setlength{\itemsep}{0.5\parsep}
        \setlength{\labelwidth}{\leftmargin}
        \setlength{\labelsep}{0em}}
    }
{\end{list}}

\providecommand{\cref}[1]{Chapter~\ref{#1}}

\newcommand{\proposed}{\textup{\textsc{SLASH}}\xspace}

\newif\ifincludesummary
\includesummaryfalse
\begin{document}

\title{Scratched Lenses, Shifted Depth: Passive Camera-Side Optical Attacks}

\author{\IEEEauthorblockN{Qinlin He, Zeming Zhuang, Yongji Wu, Lan Zhang, and Xiaoyong (Brian) Yuan}
\IEEEauthorblockA{Clemson University\\
\{qinlinh, zemingz, yongjiw, lan7, xiaoyon\}@clemson.edu}}

\IEEEoverridecommandlockouts
\makeatletter\def\@IEEEpubidpullup{6.5\baselineskip}\makeatother
\IEEEpubid{\parbox{\columnwidth}{
        Network and Distributed System Security (NDSS) Symposium 2027\\
        22--26 March 2027, Seoul, Republic of Korea\\
        ISBN 978-1-970672-09-1\\
        https://dx.doi.org/10.14722/ndss.2027.231849\\
        www.ndss-symposium.org
}
\hspace{\columnsep}\makebox[\columnwidth]{}}

\maketitle

\begin{abstract}
Physical adversarial attacks on vision systems are typically studied via scene manipulation, such as adversarial patches or projections, where the adversary controls what the camera observes at capture time. A smaller body of work has examined camera-side manipulations, such as camera stickers and auxiliary optics, which introduce consistent artifacts in the captured image. However, these approaches treat the attack as an image-space perturbation produced by designed transmissive patterns, without explicitly modeling how physical imperfections interact with scene-dependent lighting and optics. In contrast, we identify a distinct and underexplored threat: passive lens-side damage that is persistent yet trigger-conditioned, producing scene-dependent optical artifacts that bias geometric inference.

We instantiate this threat through Scratch-induced Lens Adversarial Streak Hijacking (SLASH), a physical-world attack caused by small scratches on a camera lens or protective cover. These scratches interact with bright point-like light sources and specular reflections to generate structured optical artifacts that systematically distort depth cues. Crucially, the perturbation is fixed in the optical path but conditionally triggered by the scene, making it both persistent and selective. To capture this behavior, we formulate the attack in optical space rather than pixel space, modeling the scratch pattern as a trigger-conditioned optical channel, and using the same scratch pattern across diverse viewing conditions. We evaluate SLASH in both digital and real-world settings on monocular depth estimation and monocular 3D object detection.
% \by{what is "Under the deployment-realistic fixed-scratch constraint?" Remove it if it is unnecessary.}
% \ql{RESOLVED}
\revision{%
Under the deployment-realistic fixed-scratch constraint, directional depth shifts reach up to $32\%$ relative error on monocular depth estimation, with consistent directional effects extending to monocular 3D object detection, and physical experiments confirm that the attack transfers to real camera recordings as depth shifts that substantially exceed the model's natural prediction baseline.%
}{%
On the nuScenes dataset, \proposed causes monocular depth models to estimate target vehicles as up to $32\%$ closer to the camera or $36\%$ farther away, with consistent effects extending to monocular 3D object detection.
Physical experiments with both acrylic and glass covers confirm that the attack produces substantial depth errors on real camera inputs.%
}
% \by{Can we use some numbers for the physical experiment?}
% \ql{The maximum physical shift is approximately $-60\%$. Reporting this value directly may be confusing because the physical result would appear stronger than the nuScenes result, which is counterintuitive. The physical results are also reported only in the figure rather than as exact numerical values.}
These findings reveal a new attack surface in which benign-looking hardware imperfections act as latent, scene-triggered adversarial mechanisms, challenging existing assumptions about physical robustness and highlighting the need to account for optical-path perturbations in secure vision system design.
\end{abstract}

\IEEEpeerreviewmaketitle
\setcounter{page}{1}
\begin{figure}[t]
    \centering
    \includegraphics[width=0.9\columnwidth]{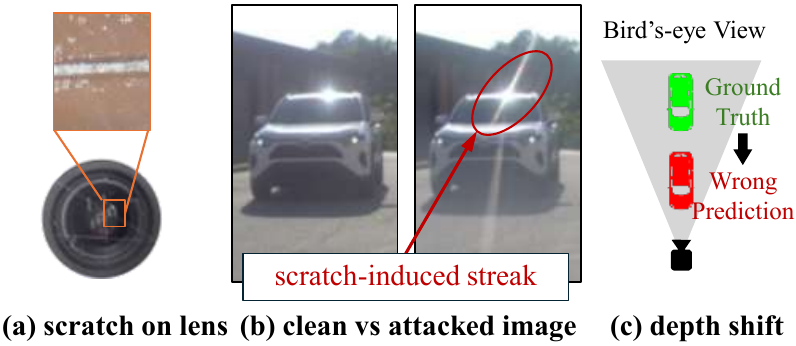}
    \vspace{-0.6em}
    \caption{\proposed Overview. (a) A small scratch on the lens or protective cover appears as ordinary wear. (b) Without a trigger (left), the image remains benign; with a compatible bright source (right), the scratch produces a streak artifact near the target. (c) The streak artifact biases the depth prediction, shifting the perceived distance of the target object.
    }
    \label{fig:intro_teaser}
    \vspace{-1.1em}
\end{figure}

\section{Introduction}
\label{sec:intro}
Adversarial attacks~\cite{goodfellow2014explaining,athalye2018synthesizing,yuan2019adversarial} have exposed broad vulnerabilities in modern vision systems, showing that carefully designed perturbations can induce model failures.
A large body of work has extended these attacks to the physical world, where the adversary perturbs the sensing process through real-world modifications rather than purely digital input changes.

Prior physical adversarial attacks on vision systems have most often manipulated either the observed scene or the imaging process at capture time. Scene-side approaches embed adversarial patches, textures, posters, or other visible artifacts in the environment. Capture-time optical attacks instead inject signals into the imaging pipeline through projected illumination, lasers, or other active optical means, including flare effects generated by controlled light sources~\cite{10806811_iot_lens_flare_attack}. In both cases, the adversary alters what the camera sees during inference, while leaving the camera itself as an unmodified observer.

A smaller line of work has shown that perturbations can also be mounted on the camera optics. The most directly comparable approach is the adversarial camera sticker, which places a translucent printed pattern on the lens to induce a universal perturbation for image classification~\cite{li2019adversarial_camera_sticker} or 2D object detection~\cite{zolfi2021_translucent_camera_sticker}. This line of work establishes the camera optics themselves as a viable attack surface, but it differs from the threat we consider in three respects. First, the sticker is a static printed pattern that overlays the same visible artifact on every captured frame regardless of scene content, making it conspicuous to ordinary visual inspection. Second, the sticker is designed against image classification or 2D object detection rather than depth estimation or 3D detection. Third, an attacker-printed sticker has no benign cover story: its presence on the lens is itself the attack, and a routine inspection that finds an unfamiliar adhesive sheet on the optics is sufficient to expose it. In contrast, our threat is geometry-targeted, mounted as a passive surface defect, and visually consistent with an ordinary lens; none of these properties is present in sticker-based attacks.
Figure~\ref{fig:intro_teaser} illustrates this threat and its effect on depth-aware perception.

We instantiate this lens-side tampering threat as \proposed (Scratch-induced Lens Adversarial Streak Hijacking), a physical-world attack realized through a small scratch on the victim camera lens or protective cover. A scratch is a natural choice for this threat: it is physically plausible, inexpensive to introduce, and subtle enough to resemble ordinary lens damage rather than a purpose-built adversarial artifact. When a bright compact light source passes through the damaged region, the scratch acts as a structured optical scatterer that produces a scratch-induced streak on the image sensor (see Figure~\ref{fig:intro_teaser}). Two properties give \proposed a stealth advantage over a printed camera sticker. First, the streak artifact is scene-triggered: the lens looks benign whenever no compatible bright source is incident on the damaged region, in contrast to a sticker that overlays a visible pattern on every captured frame. Second, the streak artifact follows the trigger source: when the trigger originates from the target vehicle (its headlights at night or specular highlights by day), the streak artifact naturally tracks the target across frames rather than remaining fixed in image space. Combined with this benign appearance, these properties make \proposed less conspicuous than a deliberately printed lens overlay.
% \by{see my revision below.}
% \ql{RESOLVED}
\revision{}{%
Both the physical scratch and the artifact it produces in the sensed image are stealthy in a deployment-relevant sense: The scratch on the camera lens is difficult to notice without close inspection of the optical front-end under favorable angle or lighting conditions, and it can appear as ordinary lens wear. When activated by a compatible bright source, the scratch may produce a streak-like artifact in the captured image. Although this artifact can be visible, it resembles a flare from a worn or dirty lens, so an observer who notices it may plausibly interpret it as a natural lens artifact instead of evidence of an attack.%
}

Such trigger conditions arise naturally in many camera-based settings whenever bright compact sources or strong specular reflections are present, including mobile imaging, robotics, and autonomous systems. Autonomous driving is a particularly important instance of this broader class because these triggers often occur near the very objects whose geometry must be inferred, making lens-side tampering especially relevant to safety-critical vision.
This attack surface is especially concerning for depth-related perception tasks, which rely on image evidence to infer geometric properties such as distance, shape, and spatial relationships.
These systems depend on fragile visual cues, including boundaries, local contrast, shading, highlights, silhouettes, and object-ground relationships, to reason about 3D structure.
Streak artifacts interfere with exactly these cues by introducing structured, geometry-inconsistent perturbations into the image.
As a result, lens-side optical tampering can distort the perceived shape and spatial layout of objects, leading to incorrect geometry predictions without altering the underlying scene.
Because such geometry estimates often serve as intermediate signals for higher-level modules, these errors can directly affect downstream perception tasks such as depth estimation and monocular 3D detection.

Realizing such an attack is technically different from conventional adversarial design: the attacker controls neither pixels nor texture directly, but instead specifies the perturbation in optical space through the geometry of a physical defect, while its image-space manifestation depends on the interaction among the defect, scene illumination, and camera imaging physics. Three technical problems follow. First, the attack must map a physical scratch to an image-plane streak artifact under camera geometry, cover distance, focus setting, and trigger location, because the artifact must arise from the optical front-end rather than be placed freely in image space. Second, the attack must synthesize a physically plausible streak artifact appearance without a full model of scratch microgeometry, lens coatings, sensor response, or image-signal processing. Third, the attack must optimize the same scratch pattern across a scene-specific frame sequence, because the scratch is fixed after deployment and cannot be retuned frame by frame.

To address these problems, \proposed treats a lens-side scratch as a constrained adversarial object in optical space, combining a geometry-constrained scratch-to-streak mapping, a lightweight appearance synthesizer, and a scene-specific depth-targeted optimization procedure. The attacker optimizes physical scratch parameters rather than pixels, and the resulting streak artifact is determined jointly by the deployed scratch, the scene-supplied trigger sources, and the camera imaging geometry.

The key novelty is a formulation in which passive optical front-end damage becomes an analyzable adversarial channel; \proposed lies outside standard pixel-space and scene-side threat models by combining camera-side placement, passive runtime operation, and fixed-after-deployment persistence. Our evaluation confirms that a small lens-side defect can materially bias monocular depth estimation and that the effect extends to monocular 3D object detection, exposing optical-path integrity as a security assumption for camera-based perception.

Our contributions are summarized as follows:
\begin{itemize}
    \item We identify passive lens-side damage as a camera-side adversarial attack surface for depth-aware perception and formalize a threat model in which a scratch-like defect is passive at runtime, persistent after deployment, and activated by scene-supplied compact light sources.

    \item We design \proposed, a physical-world attack framework that parameterizes a scratch in optical space rather than pixel space. \proposed uses a geometry-constrained scratch-to-streak-artifact mapping, a lightweight streak artifact appearance synthesizer, and scene-specific optimization to bias target depth predictions.

    \item We evaluate \proposed on monocular depth estimation and monocular 3D object detection in digital and physical experiments.
% \by{do not use 'deployment-realistic fixed-scratch constraint', use what you said in abstract}
% \ql{RESOLVED}
\revision{%
Under the deployment-realistic fixed-scratch constraint, directional depth shifts reach up to $32\%$ relative error on monocular depth estimation, with consistent directional effects extending to monocular 3D object detection; physical experiments confirm that the attack transfers to real camera recordings, producing depth shifts that substantially exceed the model's natural prediction baseline.%
}{%
On the nuScenes dataset, \proposed makes target vehicles appear up to $32\%$ closer to the camera or $36\%$ farther away in monocular depth estimation.
The directional depth shift effect also extends to monocular 3D object detection, and physical experiments show depth errors on real camera inputs that substantially exceed clean fluctuation.%
}
\revision{}{%
The evaluation also maps the boundary of the attack, covering unseen scenes and vehicles, cross-model transfer, imperfect engraving, and attack-agnostic image enhancement.%
}
\end{itemize}

% \by{no need to mention open science here.}
% \ql{RESOLVED}

\section{Related Work}

\subsection{Physical Adversarial Attacks}
\label{sec:related_physical}

Prior physical adversarial attacks against camera-based perception fall into three streams that differ in where the perturbation lives. Scene-side attacks modify the observed scene; active optical attacks inject light into the imaging process at runtime; and camera-side attacks alter the camera front-end so that artifacts are introduced before image formation completes. SLASH belongs to the third stream but instantiates a previously underexplored mechanism within it. We relate each stream to our work in turn.

\noindent\textbf{Scene-side attacks.}
% \by{don't use '--', use '-'}
% \ql{RESOLVED}
This dominant category modifies the observed scene through adversarial patches on signs or vehicles~\cite{eykholt2018_rp2_stop_sign,brown2017_adversarial_patch}, posters or textures in the environment~\cite{wang2025_adversarial_land_poster}, 3D-printed objects against camera-LiDAR fusion~\cite{cao2021_invisible_camera_lidar_fusion_attack}, and engineered shadows or reflections~\cite{zhong2022_shadow_attack,wang2023_rfla_reflective_attack}. Across this stream, the attacker controls a region of the world that the camera observes, so the attack affects any vehicle whose camera traverses the modified region. \proposed instead places the perturbation on the victim camera's optical front-end, making the attack vehicle-bound rather than location-fixed and shifting the relevant defense surface from scene-artifact detection to optical-path inspection.

\noindent\textbf{Active optical attacks.}
A second stream injects light into the imaging process at capture time, including rolling-shutter laser patterns~\cite{sayles2021_rolling_shutter}, synchronized LED arrays~\cite{guo2024_ghoststripe}, directed pulses for traffic-light recoloring~\cite{yan2022_rolling_colors_laser_traffic_light}, infrared reflections off signs~\cite{sato2024_invisible_reflections}, and projector-based ghost imagery~\cite{man2020_ghostimage}. Two works are most relevant. The DoubleStar attack~\cite{277102_double_star} directs two high-intensity sources at a stereo pair to induce a disparity error. Zhang~\etal~\cite{10806811_iot_lens_flare_attack} are closest in mechanism: they select a pre-rendered flare and optimize its image-space placement to land on a target traffic sign for misclassification. \proposed similarly exploits a light-triggered streak but engineers it through physical scratch geometry rather than selecting from a pre-rendered library, binds the trigger to the target vehicle's own illumination, and targets monocular depth and 3D detection. Unlike all active optical attacks, \proposed requires no runtime attacker presence.

\noindent\textbf{Camera-side attacks.}
A third stream modifies the camera front-end before image formation completes. The adversarial camera sticker places a translucent printed pattern on the lens to induce a universal perturbation for image classification~\cite{li2019adversarial_camera_sticker} or 2D object detection~\cite{zolfi2021_translucent_camera_sticker}. Two recent works extend this idea: Zhou~\etal~\cite{zhou2024optical} place an additional optical element in front of the camera to bias monocular depth, and Phan~\etal~\cite{phan2021_adversarial_imaging_pipeline} model the full ISP within the adversarial loop to design optics-aware perturbations. Across this stream, the perturbation is an attacker-fabricated artifact that either continuously alters every captured frame or visibly modifies the optical assembly. \proposed instead studies a passive surface defect whose adversarial artifact appears only when a compatible bright source illuminates the damaged region.

\subsection{Adversarial Attacks on Depth Estimation and 3D Detection}

Monocular depth estimation has become a core component of autonomous-driving perception~\cite{nuscenes,eigen2014_depth_multiscale,zhou2017_sfmlearner,yang2024_depth_anything_v2}, which has made it a frequent adversarial target. Existing physical attacks on this task converge on placing adversarial patches or textures in the scene. Cheng~\etal~\cite{cheng2022_physical_mde_patch} print optimized patches that, when affixed to a vehicle, produce mean depth errors above 6\,m in real driving video. Zheng~\etal~\cite{zheng2024_3d2fool_physical_mde} extend this with full 3D adversarial textures that generalize across vehicle types, viewpoints, and weather conditions, achieving errors beyond 10\,m. Zheng~\etal~\cite{299619_pi_jack} (Pi-Jack) instead manipulate geometric cues via a perspective-hijacking patch; their primary target is monocular depth, and their supplementary material additionally demonstrates the same patch transferring to 3D object detection. Liu~\etal~\cite{liu2024_advrm} (AdvRM) move the patch off the target vehicle and onto the road surface, disguising it as a road marking and exploiting the heavy reliance of MDE models on road regions. For 3D detection specifically, Wang~\etal~\cite{wang2025_adversarial_land_poster} also place physically realizable posters on the road surface to fool BEV-based detectors. All of these works share a scene-modification threat model in which the adversarial artifact is placed in the environment, and the same artifact exposes any vehicle whose camera traverses the modified region. Several of them, including Pi-Jack, already span both depth and 3D detection, so the contribution of SLASH is not the choice of tasks but the attack mechanism: SLASH targets the same downstream perception while modifying the camera hardware itself, so a deployed attack is bound to one specific vehicle rather than to a fixed location in the world.

Prior work has largely overlooked whether ordinary-looking damage to camera optics can disrupt depth-aware perception. \proposed addresses this gap by modeling how a lens-side scratch interacts with compact light sources and by optimizing the same scratch pattern across a target sequence.

\begin{figure}[!tb]
    \centering
    \includegraphics[width=0.95\linewidth]{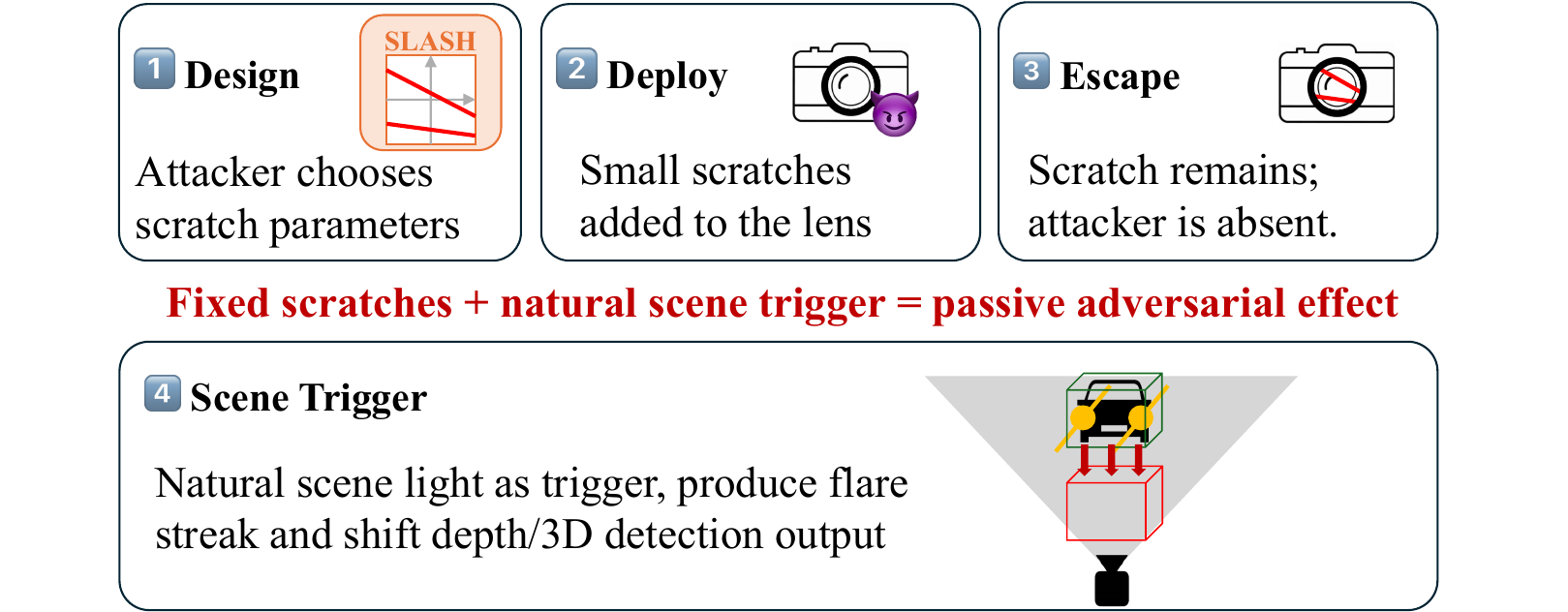}
    % \vspace{-0.5em}
    \caption{\proposed lifecycle: the attacker designs and installs a small scratch on the camera's front optical element during a window of unattended physical access; after deployment, the scratch remains fixed and the attacker is absent. When the scene provides a compatible bright source, the scratch produces a structured streak that biases depth-aware perception.}
    \label{fig:threat_model}
    \vspace{-0.6em}
\end{figure}

\section{Threat Model}
\label{sec:threat}

\newcommand{\yescirc}{\ensuremath{\bullet}}
\newcommand{\nocirc}{\ensuremath{\circ}}

\begin{table}[!tb]
\centering
\caption{Comparison between \proposed and representative physical attack classes. A filled circle (\yescirc) means the property holds, while an open circle (\nocirc) means it does not. ``Passive'' means no attacker action is required after deployment, and ``Trigger'' means condition-dependent activation.}
\label{tab:attack_positioning}
\small
\resizebox{\columnwidth}{!}{%
\begin{tabular}{lcccc}
\toprule
Attack class & Camera-side & Passive & Trigger & Persistent \\
\midrule
Scene patch/poster~\cite{eykholt2018_rp2_stop_sign,brown2017_adversarial_patch} 
  & \nocirc & \yescirc & \nocirc & \yescirc \\
Laser/projector~\cite{duan2021_adversarial_laser_beam,yan2022_rolling_colors_laser_traffic_light,man2020_ghostimage}
  & \nocirc & \nocirc & \yescirc & \nocirc \\
Adversarial lens flare~\cite{10806811_iot_lens_flare_attack} 
  & \nocirc & \nocirc & \yescirc & \nocirc \\
Camera sticker~\cite{li2019adversarial_camera_sticker,zolfi2021_translucent_camera_sticker} 
  & \yescirc & \yescirc & \nocirc & \yescirc \\
Auxiliary optical element~\cite{zhou2024optical} 
  & \yescirc & \yescirc & \nocirc & \yescirc \\
\proposed 
  & \yescirc & \yescirc & \yescirc & \yescirc \\
\bottomrule
\end{tabular}%
}
\end{table}

\noindent\textbf{Victim system and attacker goal.}
We consider camera-based perception systems that infer depth from monocular RGB images as our primary victim system. 
The primary target task is depth estimation; 3D object detection is evaluated as a representative downstream depth-aware task.
The attacker seeks to manipulate the system's depth-related prediction for a selected target object in a chosen direction, causing the system to underestimate or overestimate the object's distance relative to its true range.
For 2D depth estimation, the target object is represented by its projected image region; for 3D detection, it is represented by the matched object prediction. 
Our focus is the perception-layer attack itself rather than any specific downstream planning or control policy.

\noindent\textbf{Attacker capability and knowledge.}
The attacker obtains unattended physical access to the victim camera during a window comparable to routine parking, servicing, or inspection, and introduces one or a small number of scratches on the camera's front optical element (either the exposed lens surface or a transparent protective cover). After deployment, the scratch is fixed and the attacker is absent: they cannot adjust the perturbation online and do not compromise the camera electronics, firmware, victim software, model weights, or runtime data path.

During offline attack design, we assume a sensor-aware attacker with approximate knowledge of the camera configuration (focal length, aperture, sensor size, focus configuration) and the model family under attack. These camera parameters can be obtained from manufacturer specifications, calibration, or a surrogate setup. The optimization uses model outputs as a scoring signal from a surrogate model queried offline; it requires no gradients and no access to the deployed system after the scratch is installed.
\revision{}{%
% \by{rewrite this part - organize your writing.}
% \ql{RESOLVED}
To optimize the scratch, the attacker uses ordinary driving frames in which target vehicles appear in roughly the same image region.
These recurring target locations define the viewing geometry used for optimization and can arise across different roads and vehicles.
Table~\ref{tab:unseen_generalization} shows that scratches optimized on one set of scenes retain their effects on unseen scenes and vehicles within a similar viewing geometry.%
}

\noindent\textbf{Activation and scope.}
\proposed is trigger-conditioned: the attack activates only when a suitable bright compact source or specular reflection interacts with the scratched region and produces a streak that overlaps the target's image region. Such triggers arise naturally from target-associated headlights at night or specular highlights during the day, and need not be controlled by the attacker at runtime. The attack is therefore object-localized and deployment-constrained, and the same scratch pattern must remain effective across a bounded scene-specific frame sequence without per-frame adjustment.
\revision{}{%
% \by{Did you just mention the "fixed scratch"? Reorganize your writing.}
% \ql{RESOLVED}
The required viewing geometry and trigger condition can recur after installation.
Each recurrence can reactivate the scratch without additional attacker action.
Therefore, one installation can affect repeated encounters over the camera's service life.%
}
Figure~\ref{fig:threat_model} illustrates the attack lifecycle and Table~\ref{tab:attack_positioning} clarifies how \proposed differs from prior physical attacks.

This threat model imposes three constraints on the attack: the perturbation must be camera-side, passive at runtime, and fixed after deployment. We next explain why a small scratch satisfies these constraints and still acts as an adversarial channel against depth-aware perception.

\section{Camera-Side Optical Vulnerability}
\label{sec:insight}
Under the threat model above, the central question is why a small scratch that remains in place can affect depth-related predictions at all. Depth estimation is fundamentally underconstrained: an RGB image does not uniquely determine 3D scene geometry, so practical models rely on learned visual regularities such as object boundaries, shading transitions, highlights, silhouettes, and object-ground relationships. Because depth models process raw pixel values without distinguishing their optical origin, a local high-contrast pattern near a target object can influence the depth estimate even when that pattern is caused by the optical path rather than by the scene itself.

\noindent\textbf{Observation 1: A scratch injects structured local evidence.}
A scratch on the front optical element operates before the image reaches the perception model, behaving as a surface defect that scatters incident light anisotropically and generating an elongated streak artifact on the image plane. Because the streak is produced before digital processing, standard depth models receive it as ordinary RGB evidence rather than a marked corruption.

\noindent\textbf{Observation 2: Bright compact sources create geometry-constrained artifacts.}
The streak artifact's orientation and feasible image-plane placement are constrained by the scratch geometry, while its activation depends on the scene-supplied trigger; the attacker does not control the runtime artifact directly, but chooses the scratch pattern offline. The resulting effect manifests only when a compatible bright source illuminates the pattern. Diffuse or low-intensity illumination produces weak, spatially spread effects unlikely to influence depth predictions.

\noindent\textbf{Observation 3: The same scratch pattern must remain effective across a varying scene.}
The same scratch pattern must bias the target's depth prediction across multiple frames as the target's pose, the trigger locations, and the surrounding image context all change.
This variability motivates a scene-specific, frame-averaged optimization that jointly optimizes the pattern over a representative frame sequence.

These three observations together motivate \proposed's three-component design: a geometry-constrained scratch-to-streak mapping, a lightweight appearance synthesizer, and a scene-specific optimizer that finds a scratch pattern that remains effective across a frame sequence.

\section{SLASH Method}
\label{sec:method}

\subsection{Design Challenges and Attack Pipeline}
\label{subsec:method_challenges}

\proposed treats a scratch on the camera lens or protective cover as a physical adversarial object embedded in the optical front-end.
The attacker does not directly choose image pixels.
Instead, the attacker chooses a small number of scratch parameters on the camera lens or protective cover, and the camera image-formation process determines how the streak artifact caused by the scratch appears in captured images.
To turn such a passive defect into a depth-targeted attack, \proposed addresses three challenges.

\noindent\textbf{Challenge 1: geometry-constrained scratch-to-streak mapping.}
A scratch does not produce an arbitrary image-space streak artifact: the artifact's visible geometry is jointly constrained by scratch location, orientation, distance to the lens, focus distance, and the scene-supplied light source. The first challenge is therefore to map scratch parameters to the streak artifact's dominant geometric properties in the image plane.

\noindent\textbf{Challenge 2: physically plausible streak artifact appearance synthesis.}
Even with the streak geometry fixed, the observed intensity distribution is shaped by exposure, saturation, tone mapping, and other ISP nonlinearities to which the attacker has no access. The second challenge is therefore to synthesize a practical streak artifact appearance without relying on fully precise image formation.

\noindent\textbf{Challenge 3: scene-specific optimization.}
A deployed scratch cannot be adjusted at runtime, so the same scratch pattern must remain effective across a scene-specific frame sequence as the target pose, trigger locations, and image context evolve.

These challenges motivate the three components of \proposed. First, a geometry-constrained optical mapping computes where the scratch-induced streak appears on the image plane, which direction it follows, and how long it should be (\S\ref{subsec:geometry_mapping}). Second, a streak appearance synthesizer renders the corresponding light-like streak artifact onto the image (\S\ref{subsec:appearance_synthesis}). Third, a scene-specific depth-targeted optimization searches for a scratch pattern that biases the target's depth-related prediction across all frames in the scene (\S\ref{subsec:attack_opt}).

We use an analytic optical surrogate rather than a full differentiable path tracer such as Mitsuba3~\cite{Mitsuba3}, since a path tracer would require lens-stack, scratch microgeometry, and ISP details that are difficult to identify for a deployed camera, and is computationally expensive inside an optimization loop. Our goal is to preserve the dominant geometric constraints, not to reproduce every microscopic detail of a real scratch.

\subsection{Geometry-Constrained Scratch-to-Streak Mapping}
\label{subsec:geometry_mapping}

\begin{figure}[!t]
  \centering
  \includegraphics[width=0.65\columnwidth]{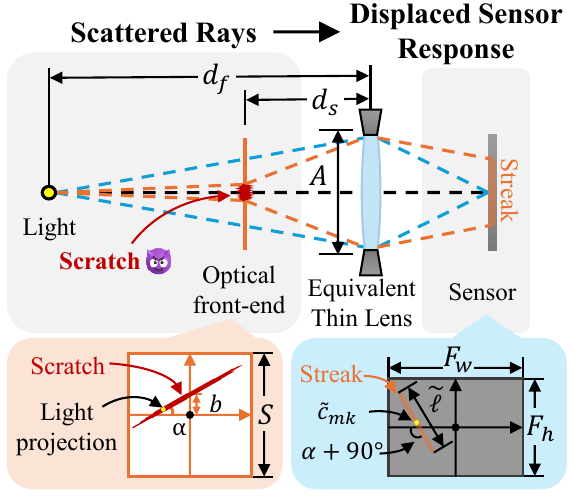}
  \vspace{-0.6em}
  \caption{Optical intuition of scratch-induced streaks. Light from a compact source follows the nominal imaging path (blue rays) through the lens to the sensor. When a scratch is present on the front optical element, part of the light is scattered (orange rays) and reaches displaced locations on the sensor, forming an elongated streak artifact of sensor-plane length $\tilde{\ell}$, centered at $\tilde{c}_{mk}$. The streak geometry is determined by the relative configuration of the scratch, lens, and focus setting; all labeled quantities ($d_f$, $d_s$, $A$, $S$, $\alpha$, $b$, $F_w$, $F_h$) are defined in the text.}
  \label{fig:light_path}
  \vspace{-0.7em}
\end{figure}

We first derive the image-plane geometry of the streak produced by a scratch on the camera front-end. A camera lens is a compound assembly, but its first-order imaging behavior can be approximated by an equivalent thin lens under the paraxial model~\cite{hecht2012optics}. The scratch lies on an accessible front optical surface, such as a protective cover, filter, or outer lens surface, located at distance \(d_s\) from this lens, while the camera is focused at distance \(d_f\), with \(d_f \gg d_s\) in normal operation.

The renderer computes three geometric quantities for each scratch-trigger pair: the streak center, orientation, and length. The center gives the clearest intuition for trigger conditioning: a bright source does not activate the entire scratch equally; the strongest artifact comes from the point on the projected scratch line closest to the trigger's image location, and the streak is centered at that point.

\noindent\textbf{Streak center.}
We model each scratch as a straight line on the front-end. This is the simplest field-realistic defect, since a single dragging motion across the cover or outer surface naturally produces a line-like scratch. Taking the front-end center as the origin of an \((x,y)\) coordinate system, we represent the \(k\)-th scratch with the Hesse normal form~\cite{bronshtein2007handbook}
\begin{equation}
x \cos\alpha_k + y \sin\alpha_k = b_k,
\label{eq:hesse-form}
\end{equation}
where \(\alpha_k\) is the angle of the scratch normal and  \(b_k \in [-0.5,0.5]\) is the normalized signed perpendicular offset from the front-end center to the scratch line, expressed relative to the side length \(S\) of the scratched front-end region. Geometrically, \(\alpha_k\) rotates the scratch line, while \(b_k\) slides it along its normal direction. With \(K\) scratches, the deployable parameters are \(\theta=\{(\alpha_k,b_k)\}_{k=1}^{K}\), giving a \(2K\)-dimensional search space that does not grow with image resolution or scene complexity.

To locate the scratch in the image, we project this front-end line onto the sensor using a paraxial thin-lens approximation. A normalized displacement on the scratched front-end maps approximately linearly to the normalized image plane by a scale factor determined by the front-end size, front-end distance, focal length, and sensor size. Let \(S\) be the side length of the scratched front-end region, \(f\) be the equivalent focal length, \(F_w \times F_h\) be the physical sensor size, and \(W \times H\) be the image resolution in pixels. In normalized image coordinates, the horizontal and vertical scale factors are given by
\begin{equation}
\kappa_u \approx \frac{S f}{d_s F_w}, \qquad \kappa_v \approx \frac{S f}{d_s F_h}.
\label{eq:magnification}
\end{equation}
The reference point on the projected scratch line, corresponding to the foot of the perpendicular from the front-end origin to the scratch, is given by
\begin{equation}
p_k = \big(0.5 - b_k \cos\alpha_k\,\kappa_u,\; 0.5 - b_k \sin\alpha_k\,\kappa_v\big).
\label{eq:scratch-reference-point}
\end{equation}
The constant \(0.5\) places the optical center at the middle of the normalized image, and the sign follows the image-coordinate convention and camera inversion.

The projected scratch-line direction in pixel coordinates is obtained by projecting a direction vector perpendicular to the normal in Eq.~\ref{eq:hesse-form}, and is given by
\begin{equation}
\widetilde{\mathbf{s}}_k = \big(W\kappa_u \sin\alpha_k,\; -H\kappa_v \cos\alpha_k\big), \qquad \hat{\mathbf{s}}_k = \frac{\widetilde{\mathbf{s}}_k}{\|\widetilde{\mathbf{s}}_k\|_2}.
\label{eq:streak-direction}
\end{equation}

Let \(L=\{(u_m,v_m)\}_{m=1}^{|L|}\) be the set of compact bright trigger locations in normalized image coordinates. In pixel coordinates, the projected scratch reference point and the trigger location are written as
\begin{equation}
q_k=(W p_{k,u}, H p_{k,v}), \qquad r_m=(u_m W, v_m H).
\end{equation}
The trigger most strongly illuminates the point on the projected scratch line nearest to its image location. We therefore define the streak center as the orthogonal projection of \(r_m\) onto the projected scratch line:
\begin{equation}
c_{mk} = q_k + \big((r_m-q_k)^\top \hat{\mathbf{s}}_k\big)\hat{\mathbf{s}}_k .
\label{eq:streak-center}
\end{equation}
This projection is the core trigger-conditioning step. The optimizer chooses where the scratch is placed on the front-end, the scene supplies the trigger location, and the optical geometry determines where the streak lands in the image. The streak center is therefore not a free image-space variable.

\noindent\textbf{Streak orientation.}
The streak is rendered along the effective projected scratch axis \(\hat{\mathbf{s}}_k\). A scratched or grooved surface scatters light anisotropically, with a dominant axis along which transmitted light preferentially spreads, so the defocused images of scattering sites along the scratch merge into an elongated artifact rather than a circular blur. The brightness profile itself is handled by the appearance surrogate in \S\ref{subsec:appearance_synthesis}.

\noindent\textbf{Streak length.}
The streak length is dominated by defocus. Since the scratch is located very close to the lens while the camera is focused on the distant scene, a point on the scratch does not form a sharp point on the sensor. Instead, it forms a large blur spot, called the circle of confusion (CoC)~\cite{hecht2012optics}. The visible streak can be viewed as a line-shaped scratch blurred by this defocus pattern. In our setting, this defocus effect dominates the characteristic spatial support of the streak.
We therefore take the CoC diameter as the characteristic support length of the rendered streak and convert it to a pixel length \(\ell_{\mathrm{px}}\); Appendix~\ref{sec:appendix_appearance} gives the derivation from focal length, f-number, focus distance, and pixel pitch.
Finite scratch length, clipping by the image boundary, and fine photometric details are handled by the appearance synthesizer below.

Once the camera and front-end are fixed, \(\ell_{\mathrm{px}}\) is constant across frames. Optimizing \(\theta\) therefore controls the streak line's orientation and reference position through scratch placement, while the per-frame trigger location determines where along this line the streak center falls, and the characteristic length is fixed by the physical camera configuration.

\subsection{Streak Appearance Synthesis}
\label{subsec:appearance_synthesis}

The mapping above determines streak center, orientation, and length. The remaining question is how the streak should look photometrically in the image plane. We do not attempt to derive a closed-form physical intensity profile. The exact profile depends on scratch microgeometry, lens coatings, sensor saturation, and the image-signal-processing pipeline, all of which are difficult to identify for a deployed camera. For our attack, this level of photometric fidelity is unnecessary. Depth-oriented models are mainly affected by the streak's location, direction, spatial support, and local contrast. We therefore use a lightweight appearance surrogate that produces a light-like streak while keeping the optimization tractable.

The surrogate rasterizes a rectangular support mask of length \(\ell_{\mathrm{px}}\) for each trigger-scratch pair, softens it with an isotropic and a directional blur, attenuates it by the trigger's perpendicular distance to the projected scratch line, and composites all pairs onto the clean image by screen blending, which brightens the streak region as extra light would rather than occluding it. Appendix~\ref{sec:appendix_appearance} gives the full formulation.

The appearance parameters are fixed during each attack run. They are calibrated once for the camera and cover setup, while the optimizer searches only over the physical scratch parameters \(\theta\). This separation keeps the attack low-dimensional while preserving the optical constraints: geometry determines where the streak appears, and the appearance surrogate determines how the light-like artifact is rendered.

\subsection{Scene-Specific Depth-Targeted Optimization}
\label{subsec:attack_opt}

After the scratch-to-streak renderer produces the attacked image, \proposed optimizes the scratch parameters to bias the target's depth-related prediction. The deployment constraint is that the scratch is fixed after installation: the attacker cannot choose a different scratch for each frame. We therefore formulate the attack as a scene-specific optimization in which the same scratch pattern is used throughout the target sequence.

Let \(\Phi(I,L;\beta,\theta)\) denote the renderer from the previous subsections, where \(I\) is the clean image, \(L\) is the set of compact bright trigger locations, \(\beta\) contains fixed camera and renderer parameters, and \(\theta\) contains the optimizable scratch parameters. The attacked image is denoted as
\begin{equation}
I' = \Phi(I,L;\beta,\theta).
\label{eq:attacked-image}
\end{equation}
The attacker selects a direction \(s_{\mathrm{dir}}\in\{+1,-1\}\), where \(s_{\mathrm{dir}}=+1\) corresponds to a \emph{closer} attack and \(s_{\mathrm{dir}}=-1\) corresponds to a \emph{farther} attack. Since the optimizer minimizes the signed loss, these two choices respectively drive the predicted target distance down or up.

\noindent\textbf{Monocular depth estimation.}
For monocular depth estimation, let \(g_{\mathrm{MDE}}\) be the depth model and
\(\hat{D} = g_{\mathrm{MDE}}(I')\)
be the predicted depth map on the attacked image. Let \(R_{\mathrm{tgt}}\) denote the 2D target region, given by the projected target bounding box. We define the predicted target depth as the mean depth inside this region:
\begin{equation}
d_{\mathrm{tgt}}
=
\frac{1}{|R_{\mathrm{tgt}}|}
\sum_{r\in R_{\mathrm{tgt}}}
\hat{D}_r.
\label{eq:mean-target-depth}
\end{equation}
The directional depth-estimation loss is defined as
\begin{equation}
\mathcal{L}_{\mathrm{MDE}}^{\mathrm{dir}}
=
s_{\mathrm{dir}}\, d_{\mathrm{tgt}}.
\label{eq:loss-mde}
\end{equation}
Thus, minimizing Eq.~\ref{eq:loss-mde} reduces the predicted target depth for the closer attack and increases it for the farther attack.

\vspace{0.2em}
\noindent\textbf{Monocular 3D object detection.}
For monocular 3D detection, let \(g_{\mathrm{DET}}\) be the detector and
\(\hat{\mathcal{B}} = g_{\mathrm{DET}}(I')\)
be the set of predicted 3D boxes. Unlike dense depth estimation, detection introduces a target-association issue: after the attack, the intended target may be shifted, confused with another object, or missing. We therefore distinguish the directional depth-shift objective from target disappearance during optimization.

We first match the attacked predictions to the intended target using the matching procedure described in Section~\ref{sec:setup}. Let \(d_{\mathrm{match}}\) be the camera-to-object distance of the matched prediction. If no valid target match exists, we assign a large positive penalty \(P\). The detection loss is denoted as
\begin{equation}
\mathcal{L}_{\mathrm{DET}}^{\mathrm{dir}}
=
\begin{cases}
s_{\mathrm{dir}}\, d_{\mathrm{match}},
& \text{if a valid target match exists,}\\
P,
& \text{otherwise.}
\end{cases}
\label{eq:loss-det}
\end{equation}
The penalty \(P\) discourages target disappearance or ambiguous association as an optimization shortcut. This forces the optimizer to search for scratches that bias the target's predicted distance while keeping the prediction associated with the intended object. Target loss is still reported as an evaluation outcome, but it is not treated as the preferred outcome for the directional optimization.

\vspace{0.2em}
\noindent\textbf{Scene-specific optimization objective.}
We define a scene-level attack instance as
\(\mathcal{E}=\{n_1,n_2,\ldots,n_N\},\)
where each frame-level sample \(n_i\) contains the image \(I_i\), trigger locations \(L_i\), target region \(R_i\), and any task-specific target context needed for matching.
\revision{%
A single-frame attack would optimize a separate \(\theta\) for each \(n_i\), but this is not physically deployable because the scratch cannot be reconfigured at runtime. Instead, \proposed optimizes the same scratch pattern over the scene:%
}{%
\proposed optimizes the same scratch pattern over the scene:%
}
\begin{equation}
\theta_{\mathcal{E}}^{*}
=
\arg\min_{\theta}
\frac{1}{N}
\sum_{i=1}^{N}
\mathcal{L}_{\mathrm{task}}
\big(n_i;\beta,\theta,s_{\mathrm{dir}}\big),
\label{eq:scene-objective}
\end{equation}
where \(\mathcal{L}_{\mathrm{task}}\) is either
\(\mathcal{L}_{\mathrm{MDE}}^{\mathrm{dir}}\) or
\(\mathcal{L}_{\mathrm{DET}}^{\mathrm{dir}}\). This objective captures the physical deployment constraint: the scratch is optimized once and remains fixed while the target pose, trigger locations, and scene context evolve over time.

The objective in Eq.~\ref{eq:scene-objective} is non-smooth in \(\theta\), as the renderer involves rasterization, clipping, and trigger-dependent attenuation, and the detection loss depends on discrete target matching. We therefore optimize \(\theta\) with a gradient-free population-based optimizer, the Covariance Matrix Adaptation Evolution Strategy (CMA-ES)~\cite{hansen2016cma}. The result is a deployable scratch pattern \(\theta_{\mathcal{E}}^{*}\) that can be physically instantiated once and evaluated across the entire scene without per-frame adjustment.

\section{Evaluation}
\revision{%
We organize the evaluation around four questions.
}{%
We evaluate \proposed across three aspects: attack effectiveness on monocular depth estimation and monocular 3D object detection, effectiveness on real camera inputs, and attack effectiveness against representative countermeasures.%
}
% \by{What are the questions? mention them explicitly, or just say "three aspects"}
% \ql{RESOLVED}
Section~\ref{sec:digital} evaluates attack effectiveness against three monocular depth estimators and three monocular 3D detectors on a curated nuScenes subset. Section~\ref{sec:physical} validates transfer to real hardware using two vertically aligned cameras mounted in a stacked configuration, enabling synchronized capture of clean and attacked views of the same scene.
Section~\ref{sec:countermeasures} evaluates the attack against representative countermeasures to characterize its resistance to defense.

\subsection{Experimental Setup}
\label{sec:setup}
\subsubsection{Digital Evaluation Setup}

\noindent\textbf{Dataset and models.}
Digital experiments use a curated subset of the nuScenes~\cite{nuscenes} validation split comprising $33$ scenes, in total $892$ annotated frames ($507$ daytime, $385$ nighttime) across the front and back cameras. Daytime and nighttime scenes are identified from scene descriptions in the nuScenes metadata.

\vspace{0.2em}
\noindent\textbf{Tasks, victim models, and attack goals.}
We evaluate on two depth-related tasks: {monocular depth estimation} (MDE) and {monocular 3D object detection} (Det3D). For each task and each frame, the attacker selects one target vehicle and one of two attack directions: \emph{closer}, which pushes the model to predict the target as nearer than its true depth, and \emph{farther}, which pushes the model to predict it as more distant. These two directions correspond to the directional objective signs in Section~\ref{subsec:attack_opt} and are optimized and reported separately throughout the evaluation. For MDE, we attack MonoDepth2~\cite{godard2019_monodepth2}, md4all~\cite{Gasperini_2023_ICCV_md4all}, and DCPI-Depth~\cite{11051129_dcpi_depth}. For Det3D, we attack FCOS3D~\cite{wang2021_fcos3d}, PGD~\cite{wang2022probabilistic_pgd}, and OVM3D~\cite{huang2024_ovm3d}; the depth target for each detection is the estimated distance from the camera to the center of the predicted vehicle bounding box, matched to the nearest ground-truth annotation under the matching protocol described in the SLASH implementation paragraph below.

\vspace{0.2em}
\noindent\textbf{Annotation procedure.}
For each frame, we manually annotate the compact bright regions on the target vehicle as the trigger locations consumed by the scratch image-formation model, paired with the existing nuScenes 3D bounding box that supplies the target ROI and the LiDAR-derived ground-truth depth; Appendix~\ref{sec:appendix_impl} gives the full procedure. Nighttime scenes are the primary evaluation target because headlights provide a stronger and more consistent trigger than daytime specular reflections; daytime results are reported for completeness.

\vspace{0.2em}
\noindent\textbf{Attack protocol.}
\revision{%
We consider two optimization protocols. The \emph{scene-specific} optimization is the main setting: the same scratch pattern is jointly optimized over all frames of a target-vehicle sequence, defined as a temporally contiguous tracklet of the same vehicle instance within a single nuScenes scene (3-20 frames, at most a 2-second gap between consecutive frames).%
}{%
The \emph{scene-specific} optimization jointly optimizes the same scratch pattern over all frames of a target-vehicle sequence, defined as a temporally contiguous tracklet of the same vehicle instance within a single nuScenes scene (3-20 frames, at most a 2-second gap between consecutive frames).%
}
This directly mirrors the deployment constraint that a physical scratch cannot be retuned per frame.
\revision{%
The \emph{single-frame} optimization relaxes this constraint by independently optimizing scratch parameters per sample; it is not physically deployable but serves as a reference upper bound that isolates the cost of using the same scratch pattern throughout the sequence. We report single-frame results alongside scene-specific results for 3D object detection.%
}{}

\vspace{0.2em}
\noindent\textbf{\proposed implementation.}
The renderer used in all experiments is the geometry-based scratch flare model of Section~\ref{subsec:geometry_mapping}, configured for the camera and lens of our physical setup.
% \by{You don't need to mention the details of these ablation studies in the main text. You just mentioned the ablation studies can be found in the Appendix, and describe the setting used in the main paper. Use the writing in revision summary.}
% \ql{RESOLVED}
\revision{%
Each attack optimizes $K=2$ scratches in joint configuration, giving a $4$-dimensional search space ($\alpha_k, b_k$ per scratch).%
}{%
Each attack uses $K=2$ scratches and jointly optimizes their $(\alpha_k,b_k)$ parameters, giving a 4-dimensional search space.
Appendix~\ref{sec:appendix_impl} provides the renderer configuration and parameter ranges, while Appendix~\ref{sec:appendix_kscratch} evaluates the choice of $K$.%
}
All main experiments use Covariance Matrix Adaptation Evolution Strategy (CMA-ES)~\cite{hansen2016cma} with a budget of $100$ objective evaluations per instance.
\revision{%
We compared CMA-ES against several alternatives in Appendix~\ref{sec:appendix_optimizer}, and found it to be the most effective within our small evaluation budget.%
}{%
Appendix~\ref{sec:appendix_optimizer} provides the remaining hyperparameters and compares alternative optimizers and evaluation budgets.%
}

For 3D object detection, scoring an attack outcome requires identifying which post-attack prediction still corresponds to the intended target; we use a geometry- and IoU-based matching protocol described in Appendix~\ref{sec:appendix_det3d_match}. When this protocol returns no valid candidate, the sample is treated as a target loss and contributes to the target loss rate (TLR) defined in Section~\ref{sec:digital} rather than to the depth-shift metric.

\vspace{0.2em}
\noindent\textbf{Evaluation metric.}
We report the \emph{relative depth error} (RE), defined as
\begin{equation}
  \mathrm{RE} = \frac{d_{\mathrm{pred}} - d_{\mathrm{gt}}}{d_{\mathrm{gt}}},
\end{equation}
where $d_{\mathrm{pred}}$ is the predicted depth and $d_{\mathrm{gt}}$ is the LiDAR-derived ground-truth depth. A positive RE means the model overestimates depth (the target appears farther than it actually is) and a negative RE means underestimation. We compute RE in two conditions per sample: $\mathrm{RE}_{\mathrm{clean}}$ when the model sees the unperturbed image, and $\mathrm{RE}_{\mathrm{attack}}$ when the model sees the same image after the SLASH renderer applies the optimized scratch pattern. The attack-induced RE change is denoted as
\begin{equation}
\Delta\mathrm{RE} = \mathrm{RE}_{\mathrm{attack}} - \mathrm{RE}_{\mathrm{clean}},
\end{equation}
which subtracts off the model's pre-existing baseline bias and therefore measures only the depth shift attributable to \proposed.

For 3D object detection, both front and back cameras are used. For monocular depth estimation, only the front camera is used for all three models. Model-specific depth calibration procedures are described in Appendix~\ref{sec:appendix_model_eval}.

\subsubsection{Physical Evaluation Setup}
The physical attack targets a single victim camera. Because outdoor recordings do not provide per-frame metric ground truth comparable to the LiDAR-derived labels in nuScenes, we obtain a clean reference by recording the same scene in parallel with a second, unmodified camera; the dual-camera setup is a measurement instrument, not part of the attack.
Two synchronized cameras carry identical removable front covers, and only one cover is engraved with the scratch pattern while the other stays clean as the control. Appendix~\ref{sec:appendix_physical_setup} details the camera setup and why the scratch is placed on the cover rather than on the bare lens.

\begin{figure}[t]
  \centering
  \begin{subfigure}[t]{0.3\columnwidth}
    \centering
    \includegraphics[width=\linewidth]{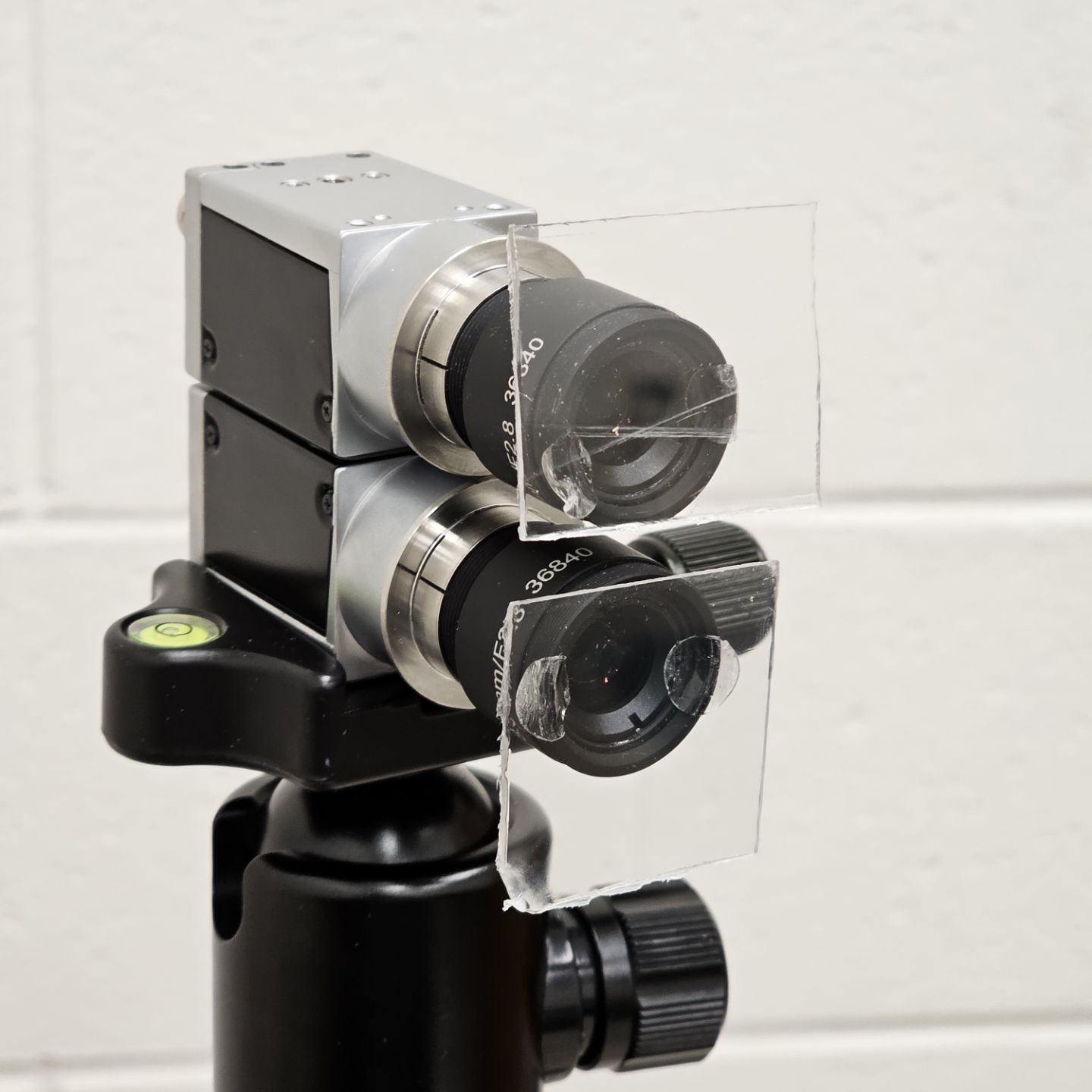}
    \caption{Camera setup.}
  \end{subfigure}
  \hfill
  \begin{subfigure}[t]{0.3\columnwidth}
    \centering
    \includegraphics[width=\linewidth]{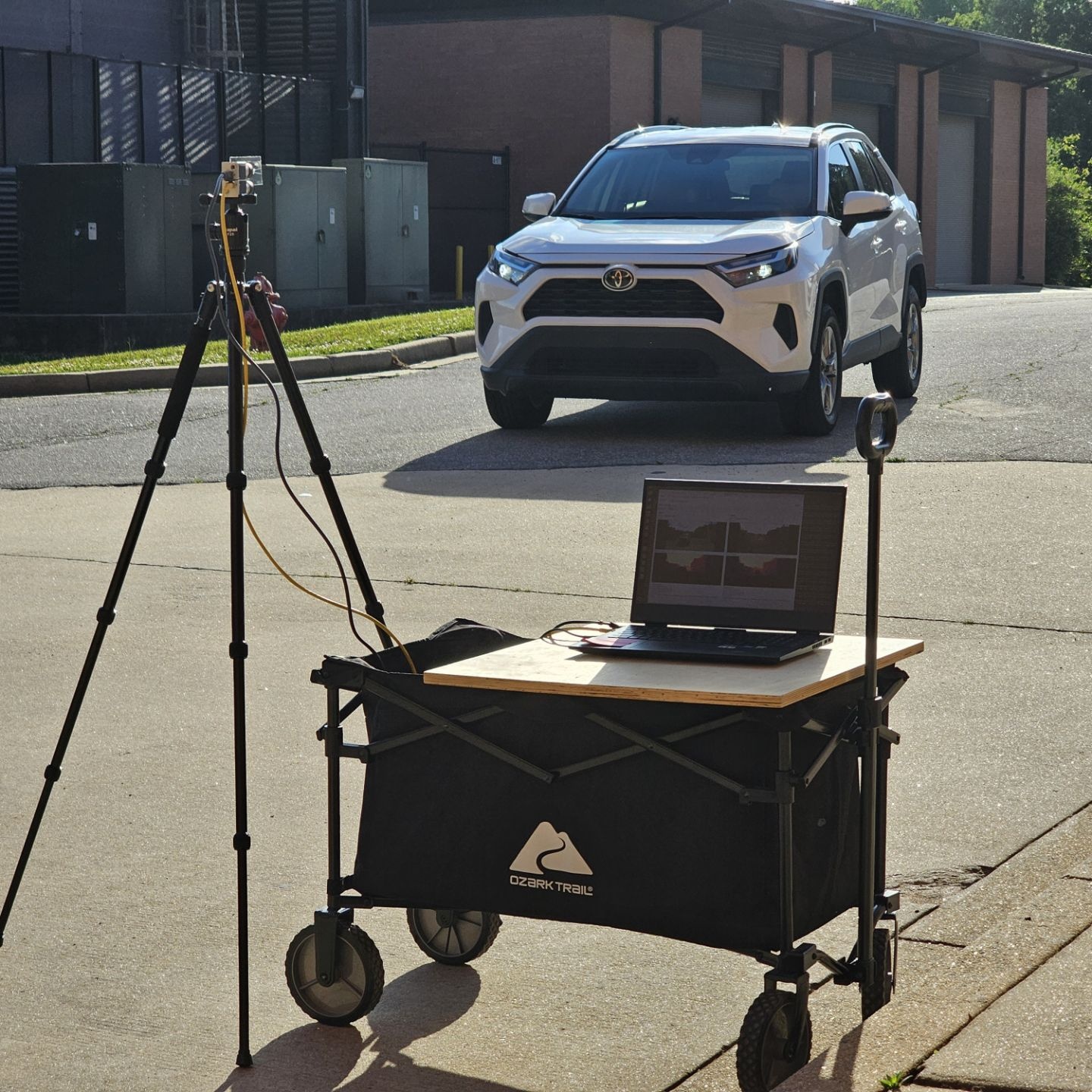}
    \caption{Test site.}
  \end{subfigure}
  \hfill
  \begin{subfigure}[t]{0.3\columnwidth}
    \centering
    \includegraphics[width=\linewidth]{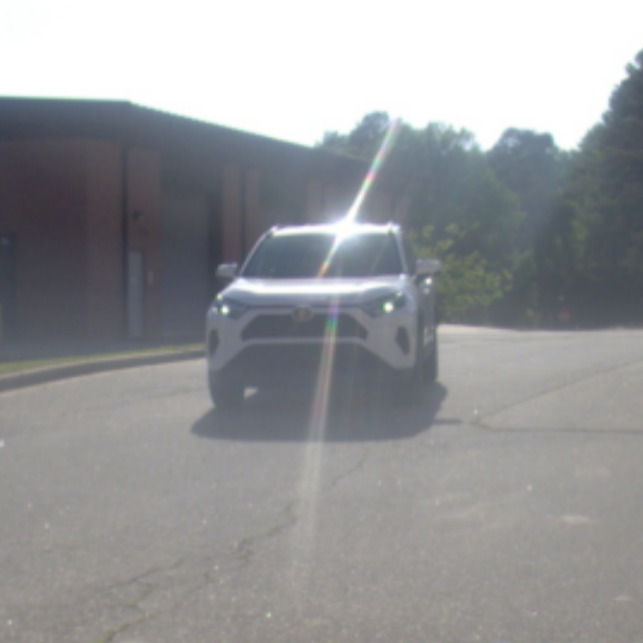}
    \caption{Camera capture.}
  \end{subfigure}
  \caption{Physical experiment setup. (a) Dual-camera close-up with clean and scratched acrylic covers; one cover is clean and serves as the clean control, while the other is engraved with the attack scratch pattern. (b) Overall test site, with the white SUV target and the tripod-mounted cameras. (c) Frame captured by the attacked camera, showing the scratch-induced streak that the depth model receives as input.}
  \label{fig:physical_setup}\vspace{-1.1em}
\end{figure}

% \by{this sentence is hard to follow. The reviewer may not be able to get that acrylic is a type of polymer. Make your writing consistent, just use acrylic and glass. No other terms. Just mention the cover is made by two materials: acrylic and glass. Reader cannot get microscope slide is equivalent to glass. Put the detailed steps in the appendix.}
% \ql{RESOLVED}
\revision{}{%
We conduct the physical experiments with both acrylic and glass covers of the same thickness.
For either material, the same scratch pattern is used throughout each recording after optimization on nuScenes.
Figure~\ref{fig:physical_setup}(a) shows the acrylic covers mounted on the cameras, while Figure~\ref{fig:glass_fabrication} shows the glass cover used in our experiments and the tool used to engrave it.
Results for acrylic and glass covers are reported throughout Section~\ref{sec:phys_dist_angle} and Section~\ref{sec:phys_scenes}.%
}

Figure~\ref{fig:physical_setup} shows both the camera setup and the overall test site, in an open area adjacent to a parking lot, with a white SUV used as the target vehicle. Because LiDAR ground truth is unavailable in this setting, we treat the synchronized clean view's prediction as the surrogate ground-truth depth and reuse the digital RE / $\Delta\mathrm{RE}$ metric directly (Section~\ref{sec:phys_dist_angle} discusses this substitution). For OVM3D, we extract the nearest detected vehicle depth and additionally report target loss rate (TLR); for md4all, we rerun the model on the recorded RGB frames and average the predicted depth inside the annotated target ROI. We conduct two complementary physical studies: an angular effectiveness characterization using OVM3D (Section~\ref{sec:phys_dist_angle}), in which the vehicle is held still while the camera viewing angle is varied; and an approaching-vehicle evaluation using md4all and OVM3D (Section~\ref{sec:phys_scenes}), in which the cameras are static but the target vehicle drives toward the camera, traversing a wide depth range under realistic motion.

\subsection{Digital Experiments}
\label{sec:digital}

We first evaluate \proposed on monocular depth estimation, the primary target task, and then assess whether the same camera-side optical perturbation affects monocular 3D object detection.
\revision{%
In both tasks, the main setting is \emph{scene-specific} optimization, where the same scratch pattern is used throughout the target sequence without per-frame adjustments.%
}{%
In both tasks, we use \emph{scene-specific} optimization, where the same scratch pattern is used throughout the target sequence without per-frame adjustments.%
}
\revision{%
For 3D detection, we also report a \emph{single-frame} setting, in which scratch parameters are optimized for each frame.%
}{}

\subsubsection{Attack against Monocular Depth Estimation}

Monocular depth estimation is the most direct target for \proposed because the attack objective is defined in terms of the predicted depth of the target region.
Table~\ref{tab:monodepth} shows that using the same scratch pattern throughout a target sequence can induce large directional errors.
Across the three models, the attack is effective in both closer and farther directions, although the strongest direction depends on the model. MonoDepth2 and md4all are most affected in the closer direction, reaching \(\Delta\mathrm{RE}=-29.12\%\) and \(-32.43\%\) at night, respectively. DCPI-Depth shows the strongest farther-direction response, reaching \(+35.90\%\) at night.

The day/night pattern is also model-dependent. md4all and DCPI-Depth show their strongest effects at night, consistent with bright headlights acting as compact triggers. MonoDepth2 is more mixed: its closer attack is stronger at night, while its farther attack is stronger during the day. This indicates that the scratch artifact is not simply a generic brightness corruption; rather, it interacts with each depth model's learned cues and scale behavior. For DCPI-Depth, which is scale-ambiguous, we use the clean-derived target-local scaling described in Section~\ref{sec:setup}, so the attacked prediction is not rescaled in a way that would absorb the attack effect.

% \by{Just use "Rand column"}
% \ql{RESOLVED}
\revision{}{%
The random-scratch control isolates the contribution of optimization.
The Rand column reports the mean effect of random scratch patterns sampled from the same parameter bounds without optimization.
In every condition, random scratches cause a smaller-magnitude shift than both optimized directions.
Thus, on the model against which it is optimized, optimization produces a larger and more consistently directed shift than a random scratch.%
}

% \by{What is "8"?}
% \ql{The "8" refers to the combinations of two models, two lighting conditions, and two attack directions. The result is reported as the maximum retained effect without relying on this count.}
\revision{}{%
Because the scratch must be engraved before the victim camera encounters a particular vehicle, it must remain effective beyond the data used for optimization.
To test this requirement, we group target samples into bins by distance and azimuth angle and assign entire scenes within each bin to either the optimization or evaluation split, so every evaluation sample contains a scene and target vehicle unseen during optimization.
Table~\ref{tab:unseen_generalization} shows that the best result retains approximately $90\%$ of the in-sample effect on these unseen scenes and vehicles.
This generalization applies when the target's distance and azimuth angle are similar to those represented in the optimization split and the scratch-induced streak overlaps the target region.%
}

\begin{figure*}[t]
  \centering
  \begin{tikzpicture}
    \node[inner sep=0pt, outer sep=0pt] (img) at (0,0) {%
      \includegraphics[width=0.80\textwidth]{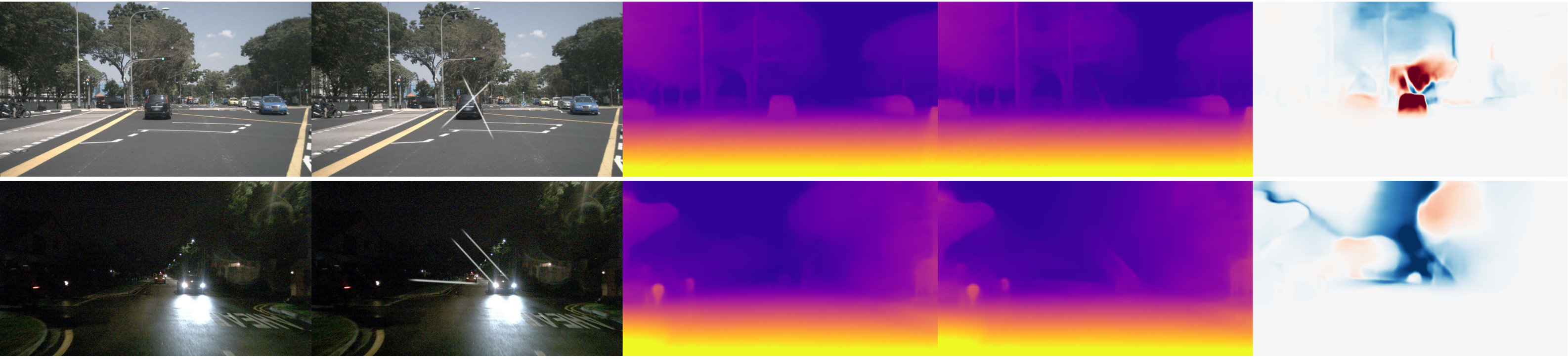}%
    };
    \begin{scope}[every node/.style={anchor=north, font=\small, yshift=-1mm}]
        \node at ($(img.south west)!0.10!(img.south east)$) {Clean};
        \node at ($(img.south west)!0.30!(img.south east)$) {Attacked};
        \node at ($(img.south west)!0.50!(img.south east)$) {Clean Depth};
        \node at ($(img.south west)!0.70!(img.south east)$) {Attacked Depth};
        \node at ($(img.south west)!0.90!(img.south east)$) {Depth Difference};
    \end{scope}
    \node[rotate=90, anchor=south, font=\small] at ($(img.west)!0.5!(img.north west)$) {Day};
    \node[rotate=90, anchor=south, font=\small] at ($(img.west)!0.5!(img.south west)$) {Night};
  \end{tikzpicture}
  \vspace{-0.6em}
  \caption{Scene-specific attack examples on monocular depth estimation using md4all. Each row shows one target vehicle sample, and the columns show the clean RGB image, the attacked RGB image, the clean depth prediction, the attacked depth prediction, and the attacked-clean depth difference. The examples illustrate that \proposed can alter the target's local depth prediction and nearby depth structure.}
  \label{fig:md4all_per_sequence_qualitative}\vspace{-1.1em}
\end{figure*}

Figure~\ref{fig:md4all_per_sequence_qualitative} gives representative md4all examples. The attacked images show localized streak artifacts near the target, and the corresponding depth maps show that these local artifacts can change both the target region and the nearby depth structure.

\revision{}{%
In a separate MonoDepth2 nighttime farther-direction sweep, the default brightness of $1.0\times$ and width of $0.0075$ produce a $+10.43\%$ depth shift at $28.50$\,dB peak signal-to-noise ratio (PSNR).
Increasing brightness to $4.0\times$ raises the shift to $+13.31\%$ but lowers PSNR to $21.84$\,dB; increasing width to $0.015$ raises the shift to $+12.81\%$ and lowers PSNR to $26.38$\,dB.
Thus, brighter or wider rendered streaks can strengthen the attack at the cost of image-level stealthiness.%
}

\begin{table}[t]
  \centering
  \caption{\revision{%
Scene-specific attack results on monocular depth estimation. A single scratch is jointly optimized over all frames of the same target sequence. C and F denote closer and farther attack directions. Several settings exceed \(20\%\) target-depth shift, showing that dense depth prediction is highly sensitive to scratch-induced streak artifacts.%
}{%
Scene-specific attack results on monocular depth estimation. The same scratch pattern is used throughout each target sequence. The Rand column reports the mean effect of random scratch patterns sampled from the same parameter bounds. Clean RE and $\Delta$RE are reported as percentages.%
}}
  % \by{remove std from Rand.}
  % \ql{RESOLVED}
  \label{tab:monodepth}
  \small
  \resizebox{0.95\columnwidth}{!}{%
  \begin{tabular}{ll S[table-format=+2.2] c S[table-format=+2.2] S[table-format=+2.2]}
    \toprule
    \multirow{2}{*}{Model} & \multirow{2}{*}{Scene} & \multicolumn{1}{c}{\multirow{2}{*}{Clean RE}} & \multicolumn{3}{c}{$\Delta$RE} \\
    \cmidrule(lr){4-6}
    & & & \revision{}{Rand} & \multicolumn{1}{c}{Closer} & \multicolumn{1}{c}{Farther} \\
    \midrule
    \multirow{2}{*}{MonoDepth2}
      & Day   & +9.56  & \revision{}{\tablenum[table-format=+1.2]{-3.04}} & -20.34 & +14.32 \\
      & Night & -11.23 & \revision{}{\tablenum[table-format=+1.2]{-7.81}} & -29.12 & +8.52 \\
    \midrule
    \multirow{2}{*}{md4all}
      & Day   & +6.59  & \revision{}{\tablenum[table-format=+1.2]{-1.12}} & -15.74 & +11.53 \\
      & Night & +13.92 & \revision{}{\tablenum[table-format=+1.2]{-3.30}} & -32.43 & +23.00 \\
    \midrule
    \multirow{2}{*}{DCPI-Depth}
      & Day   & -0.36 & \revision{}{\tablenum[table-format=+1.2]{-0.64}} & -9.31 & +4.99 \\
      & Night & +2.87 & \revision{}{\tablenum[table-format=+1.2]{+6.20}} & -8.84 & +35.90 \\
    \bottomrule
  \end{tabular}%
  }
  \vspace{-0.5em}
\end{table}

\begin{table}[t]
  \centering
  \caption{\revision{}{%
% \by{make 'Retained' clear, e.g. retained of depth shift}
% \ql{RESOLVED}
Generalization to unseen scenes and vehicles. Target vehicles are grouped into bins by distance and azimuth angle. Within each bin, a scratch is optimized on the optimization split and evaluated on the evaluation split, which contains unseen scenes and vehicles. The Seen column reports $\Delta\mathrm{RE}$ (\%) on the optimization split, while the Unseen column reports it on the evaluation split. Retained is the depth shift on unseen scenes and vehicles divided by the depth shift on the optimization split (Unseen divided by Seen).%
% \by{no DCPI-depth result?}
% \ql{RESOLVED}
}}
  \label{tab:unseen_generalization}
  \small
  \setlength{\tabcolsep}{3pt}
  \begin{tabular}{llc S[table-format=+2.2] S[table-format=+2.2] c}
    \toprule
    Model & Time & Direction & \multicolumn{1}{c}{Seen} & \multicolumn{1}{c}{Unseen} & Retained \\
    \midrule
    \multirow{4}{*}{MonoDepth2}
      & \multirow{2}{*}{Day}   & Closer  & -20.95 & -13.02 & $62\%$ \\
      &                        & Farther & +13.42 & +6.14  & $46\%$ \\
      & \multirow{2}{*}{Night} & Closer  & -28.76 & -25.13 & $87\%$ \\
      &                        & Farther & +9.19  & +2.92  & $32\%$ \\
    \midrule
    \multirow{4}{*}{md4all}
      & \multirow{2}{*}{Day}   & Closer  & -15.42 & -3.62  & $24\%$ \\
      &                        & Farther & +10.07 & +5.28  & $52\%$ \\
      & \multirow{2}{*}{Night} & Closer  & -27.99 & -12.08 & $43\%$ \\
      &                        & Farther & +31.19 & +28.15 & $90\%$ \\
    \midrule
    \multirow{4}{*}{DCPI-Depth}
      & \multirow{2}{*}{Day}   & Closer  & -8.84  & -6.08  & $69\%$ \\
      &                        & Farther & +4.68  & +3.21  & $69\%$ \\
      & \multirow{2}{*}{Night} & Closer  & -6.87  & -6.32  & $92\%$ \\
      &                        & Farther & +35.30 & +22.22 & $63\%$ \\
    \bottomrule
  \end{tabular}
\end{table}

\subsubsection{Attack against 3D Object Detection}

% \by{reorganize the paragraphs here. merge some paragraphs. Each paragraph speaks for one thing.}
% \ql{RESOLVED}
We next evaluate monocular 3D object detection, where an attack can shift the estimated distance of a detected target or cause the target to no longer be detected.
Therefore, Table~\ref{tab:det3d_persequence} reports both directional depth shift and target loss rate (TLR).
For detected targets, the strongest effects occur at night in the farther direction, where $\Delta\mathrm{RE}$ reaches $+6.46\%$ for FCOS3D, $+6.15\%$ for PGD, and $+8.51\%$ for OVM3D.
The closer direction is weaker overall, with OVM3D reaching $-7.52\%$ at night, while daytime attacks produce smaller shifts across all three detectors.
\revision{%
Figure~\ref{fig:ovm3d_per_sequence_qualitative} shows representative OVM3D examples in which the target vehicle remains detected but shifts in depth.%
}{}

TLR captures a separate failure mode in which a target detected in the clean image is no longer detected after the attack. These samples are excluded from $\Delta\mathrm{RE}$ because no valid target detection remains for computing a depth shift. Daytime attacks cause $5\%$ to $8\%$ of clean-detected targets to be lost. At night, TLR remains below $1\%$, indicating that the attack primarily changes the estimated distance while preserving the target detection.

% \by{use 'from .. to ..' for values with signs, '-+' looks weird}
% \ql{RESOLVED}
% \by{use '-' instead of '--'}
% \ql{RESOLVED}
\revision{}{%
We further compare optimized and random scratch patterns using $\Delta\mathrm{RE}$.%
Random patterns keep $\Delta\mathrm{RE}$ within $\pm 1.3\%$ across all detectors and conditions, indicating that optimization produces larger directional depth shifts.%
}

Overall, the digital experiments show that \proposed induces directional depth errors under scene-specific optimization. Dense depth models exhibit the largest shifts, while 3D detectors show smaller but consistent distance errors and occasional target loss. These results motivate the physical study below, which tests whether the scratch-induced artifacts transfer to real camera inputs.

\begin{table}[t]
  \centering
  \caption{\revision{%
Scene-specific attack results on 3D object detection. A single scratch is jointly optimized over all frames in which the same vehicle instance is tracked. C and F denote closer and farther attack directions. TLR is the fraction of clean-detected targets that become undetectable or unmatchable after attack, averaged over the two attack directions.%
}{%
Scene-specific attack results on 3D object detection. The same scratch pattern is used throughout each target sequence. Rand reports the mean $\Delta\mathrm{RE}$ across random scratch patterns. TLR is the optimized attack's target loss rate averaged across the closer and farther attacks. Clean RE, $\Delta$RE, and TLR are reported as percentages.%
}}
  \label{tab:det3d_persequence}
  \small
  \resizebox{\columnwidth}{!}{%
  \begin{tabular}{ll S[table-format=+1.2] c S[table-format=+1.2] S[table-format=+1.2] S[table-format=2.2]}
    \toprule
    \multirow{2}{*}{Model} & \multirow{2}{*}{Scene} & \multicolumn{1}{c}{\multirow{2}{*}{Clean RE}} & \multicolumn{3}{c}{$\Delta$RE} & \multicolumn{1}{c}{\multirow{2}{*}{TLR}} \\
    \cmidrule(lr){4-6}
    & & & \revision{}{Rand} & \multicolumn{1}{c}{Closer} & \multicolumn{1}{c}{Farther} & \\
    \midrule
    \multirow{2}{*}{FCOS3D}
      & Day   & +2.65 & \revision{}{\tablenum[table-format=+1.2]{+0.20}} & -2.06 & +3.39 & 8.06 \\
      & Night & -0.33 & \revision{}{\tablenum[table-format=+1.2]{+1.28}} & -1.65 & +6.46 & 0.21 \\
    \midrule
    \multirow{2}{*}{PGD}
      & Day   & +2.18 & \revision{}{\tablenum[table-format=+1.2]{+0.04}} & -2.05 & +2.87 & 5.76 \\
      & Night & -0.84 & \revision{}{\tablenum[table-format=+1.2]{+0.78}} & -1.93 & +6.15 & 0.75 \\
    \midrule
    \multirow{2}{*}{OVM3D}
      & Day   & -4.04 & \revision{}{\tablenum[table-format=+1.2]{-0.51}} & -4.33 & +3.40 & 6.73 \\
      & Night & -7.16 & \revision{}{\tablenum[table-format=+1.2]{-0.41}} & -7.52 & +8.51 & 0.77 \\
    \bottomrule
  \end{tabular}%
  }\vspace{-0.5em}
\end{table}

\subsection{Physical Experiments}
\label{sec:physical}

\revision{%
Physical experiments validate that the attack transfers beyond simulation and clarify when it is strongest in real scenes.%
}{%
Physical experiments evaluate whether scratch patterns optimized on nuScenes remain effective on real camera inputs.
The scratch patterns are engraved on the physical covers without optimization using data from the physical experiments.%
}
We organize the study into two complementary parts. We first hold the target vehicle stationary and evaluate the attack with the camera placed at several fixed viewpoints, which allows us to measure whether \proposed produces a measurable depth bias on real hardware and to characterize how that bias varies across viewing configurations (Section~\ref{sec:phys_dist_angle}). We then hold the cameras stationary and let the target vehicle drive toward them, exposing the attack to a continuous depth-range traversal under realistic motion (Section~\ref{sec:phys_scenes}).

\subsubsection{Attack Effectiveness Across Camera Viewing Angles}
\label{sec:phys_dist_angle}
We first validate that \proposed transfers to physical hardware: we want to know whether a deployed lens scratch produces a measurable depth bias on real frames, not just on the nuScenes dataset. The static camera setup described in Section~\ref{sec:setup} is well suited to this question because it pinpoints all confounds except the optical path, so any systematic difference between the clean and attacked recordings can be attributed to the scratch itself.

Concretely, we record paired clean and attacked OVM3D streams while holding both the cameras and the target vehicle stationary and varying the dominant illumination angle. Because no per-frame LiDAR ground truth is available outdoors, we treat the synchronized clean view's prediction as the surrogate ground-truth depth and reuse the RE / $\Delta\mathrm{RE}$ metric of Section~\ref{sec:setup}: $d_{\mathrm{gt}}$ is the nearest detected vehicle depth in the clean view, $d_{\mathrm{pred}}$ is the same quantity in the synchronized attacked view, and $\Delta\mathrm{RE}$ then captures the depth shift attributable to the scratched optics. An angle of $0^\circ$ denotes near-direct alignment between the dominant light source and the camera optical axis.
\revision{%
To distinguish attack-induced shifts from ordinary prediction noise, we additionally report a \emph{clean fluctuation baseline} computed from the clean-view trajectory alone: for each frame, we fit a short-window local trend to the clean depth sequence and measure the residual deviation from that trend, capturing the model's natural frame-to-frame variation in the absence of any attack.%
}{%
To distinguish attack-induced shifts from ordinary prediction noise and the systematic offset due to slight differences between the two cameras, we compare each curve against a dual-clean reference recorded with an unmodified cover on both cameras.
The mean of that reference is the residual difference between the two cameras in the absence of scratches.
Around this mean, we draw a baseline band whose half-width is one clean fluctuation, obtained by fitting a short-window local trend to the clean depth sequence and measuring the residual deviation from that trend.
This deviation captures the model's frame-to-frame variation under attack-free conditions.
Therefore, a prediction that remains within the band is likely due to ordinary variation, while one that falls outside it is likely due to the attack.%
}

\begin{figure}[!tb]
  \centering
  \includegraphics[width=0.9\columnwidth]{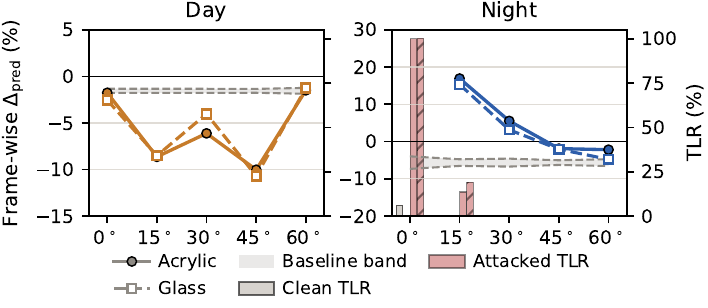}
  \caption{%
\revision{%
Physical OVM3D angle sweep.%
}{%
Attack effectiveness at different viewing angles.%
}
% \by{Attack effectiveness at different viewing angles. }
% \ql{all image/table captions now convey the core idea that we want to convey via this figure/table in the first sentence}
% \by{remove caption in the figure (Physical OVM3D angle sweep)}
% \ql{RESOLVED}
\revision{%
The left y-axis shows the attacked vs. clean relative change in nearest-vehicle depth as colored lines, while the gray band denotes the clean fluctuation baseline derived from the clean-view trajectory alone.
The right y-axis reports clean and attacked TLR.%
}{%
The left y-axis shows the relative change in target vehicle depth (attacked vs. clean) as colored lines, while the gray band is the baseline.
The right y-axis reports clean and attacked TLR.%
}
\revision{}{%
Solid lines and solid bars are the acrylic cover.
Dashed lines and hatched bars are the glass cover.%
}}
  \label{fig:phys_angle}\vspace{-1.1em}
\end{figure}

% \by{revise this paragarph, it's too long and has too much detail, no need for that much details, the conclusion is more important than the details}
% \ql{RESOLVED}
Figure~\ref{fig:phys_angle} shows that the physical effect is not only measurable but substantially larger than ordinary prediction noise under attack-favorable conditions, confirming that \proposed transfers to real hardware.
\revision{%
During the day, the attack remains effective over multiple illumination angles, with the largest farther bias at $15^\circ$ ($+13.0\%$) and another clear peak at $45^\circ$ ($+9.3\%$), while the clean fluctuation baseline stays below $1\%$ and both clean and attacked TLR remain at $0\%$.%
}{%
During the day, the attack produces two substantial depth-shift peaks at $15^\circ$ and $45^\circ$, including a $-10.7\%$ shift, while both clean and attacked TLR remain at $0\%$ across the sweep.
These two peaks occur because sunlight reflected from the vehicle body can trigger the scratch at multiple angles.
At the remaining angles, the streak forms weakly and the shift approaches the baseline band.%
}
\revision{%
At night, the effect becomes even stronger but more directional: at $0^\circ$ the attacked camera loses the target vehicle in all frames (attacked TLR $100\%$, versus clean TLR $6.1\%$); at $15^\circ$ it still causes a large farther shift of $+16.9\%$ with a clean fluctuation baseline of only $0.9\%$ together with $13.5\%$ attacked TLR; and by $45^\circ$--$60^\circ$ the shift approaches the natural fluctuation range.%
}{%
At night, the depth shift at $15^\circ$ reaches $+16.9\%$, well outside the baseline band.
The directional headlight beam confines this stronger nighttime effect to a narrower angular window.
At direct alignment, the attacked TLR reaches $100\%$, compared with $6.1\%$ for the clean camera.
This difference arises because the directly aligned headlights and scratch-induced streak exceed the camera's dynamic range, saturating the target region and producing bloom that prevents detection.
This mechanism is consistent with the high-beam setting, where extreme brightness contrast similarly saturates the target region and produces bloom, causing both clean and attacked cameras to reach $100\%$ TLR.%
}
\revision{%
This day/night asymmetry is consistent with the underlying physics: the nighttime streak artifact is dominated by the target headlights and is strongest near direct alignment, whereas the daytime streak artifact can also be driven by sunlight reflected from the vehicle body, supplying multiple effective angles.%
}{}
% \by{what is source intensity?}
% \ql{Replaced "source" to "light source", "intensity" to "brightness"}
\revision{%
The nighttime TLR behavior is especially important because it shows that the scratched optics can induce a full representation failure, not just a small metric perturbation.%
}{%
Across the day and night sweeps, acrylic and glass covers produce similar trends.
Overall, compatible illumination produces scratch-specific depth shifts or target loss beyond ordinary variation, while saturation and bloom under extreme brightness can independently increase TLR.%
}

\subsubsection{Attack Effectiveness Over a Continuous Depth Range}
\label{sec:phys_scenes}

In the viewpoint study above, the target vehicle was held still while we varied the illumination angle; we now evaluate the attack under realistic motion by letting the target vehicle drive toward the still-stationary cameras over time, traversing a wide range of depths within a single recording. For md4all, we rerun the model on every recorded RGB frame and compute the mean predicted depth inside the annotated ROI. For OVM3D, we again use the nearest detected vehicle depth. Because OVM3D sometimes fails to detect the vehicle on the clean stream before it enters the reliable detection range, TLR is computed only within the temporal window between the first and last clean detections of the target vehicle.

\begin{figure}[t]
  \centering
  \includegraphics[width=0.8\columnwidth]{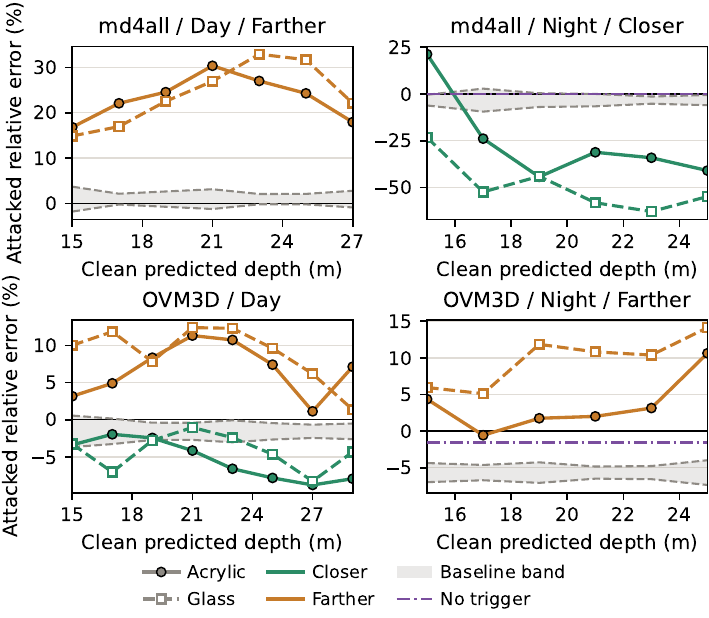}
  \caption{
    \revision{%
      Physical approaching-vehicle-scene summary using 2.5\,m clean-depth bins.%
    }{%
      Attack effectiveness at different target vehicle distances.
      Results are grouped using a shared 2.0\,m clean-depth bin width.%
    }
    \revision{%
      Colored lines show the attacked vs. clean relative change within each depth bin, and the gray band shows the clean fluctuation baseline.%
    }{%
      Colored lines show the relative change in attacked vs. clean within each depth bin, while the gray band is the baseline.%
    }
    \revision{}{%
      Solid lines are the acrylic cover.
      Dashed lines are the glass cover.
      The dash-dotted line in the two nighttime panels is the no-trigger control.%
    }
  }
  \label{fig:phys_forward}
  \vspace{-1em}
\end{figure}

% \by{The result below is new? }
% \ql{The acrylic results are old, the glass results are new.}
Figure~\ref{fig:phys_forward} summarizes the approaching-vehicle-scene results using 2.0\,m clean-depth bins.
Rather than collapsing each recording into a single average, we analyze the attack by depth range and group trajectories according to their local behavior in the effective region shown in the figure.
\revision{%
Each panel also includes a clean fluctuation baseline computed from the clean-view trajectory after removing the local depth trend, so the reader can directly compare attack-induced shifts against the model's ordinary variation in the same depth regime.%
}{%
Each panel also includes a baseline band, whose half-width is computed per depth bin from the clean-view trajectory after removing the local depth trend, so attack-induced shifts can be compared against the two controls in the same depth regime.%
}
This view is more informative for approaching-vehicle scenes than a single per-recording average because the target traverses a wide depth range and the attack is not uniformly strong throughout the whole sequence.

% \by{you never mention what is transferred attack...}
% \ql{"transfer" at here means the scratch pattern is reused from the nuScenes optimization results, I have adjusted the description}
The attack is strongest at intermediate distances as the target vehicle approaches the camera.
The effective range lies roughly between 15\,m and 30\,m clean predicted depth, with the strongest distortion appearing around 20\,m.
Within this range, the attack-induced shift exceeds the baseline band in most situations. In the strongest case (md4all during the day in the farther direction), the relative shift approaches $30\%$ around 20\,m, corresponding to a prediction moving from roughly 20\,m in the clean view to roughly 26\,m in the attacked view.

\begin{figure}[t]
  \centering
  \begin{subfigure}[t]{\columnwidth}
    \centering
    \includegraphics[width=0.95\linewidth]{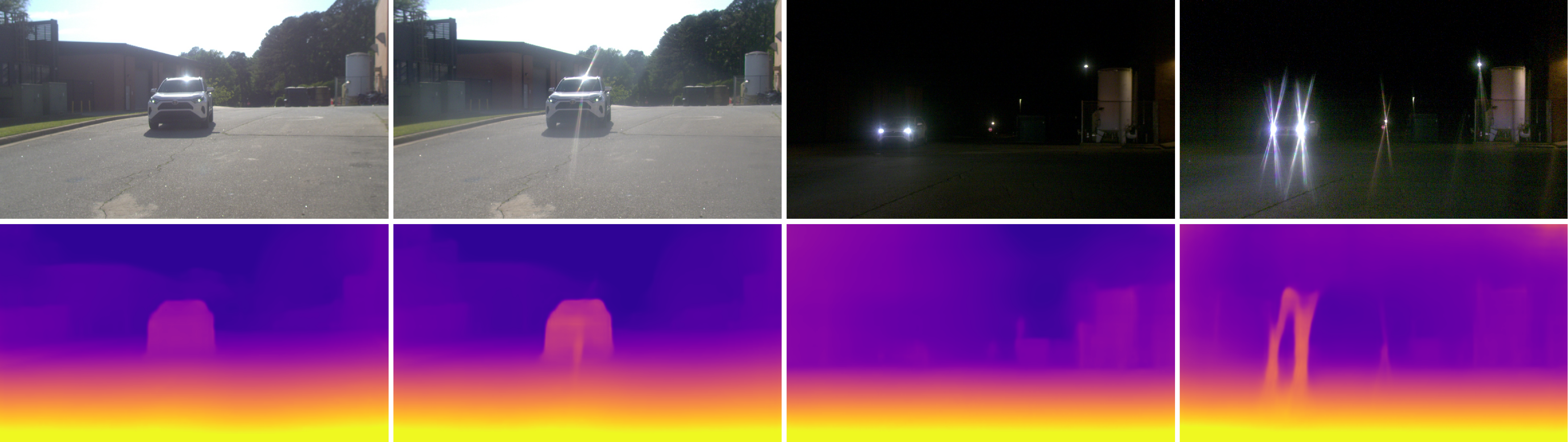}
    \caption{Representative md4all physical examples in daytime and nighttime approaching-vehicle scenes. For each sample, the top row shows the clean and attacked RGB images, and the bottom row shows the corresponding clean and attacked depth predictions.}
  \end{subfigure}
  \par\vspace{0.3em}
  \begin{subfigure}[t]{\columnwidth}
    \centering
    \includegraphics[width=0.85\linewidth]{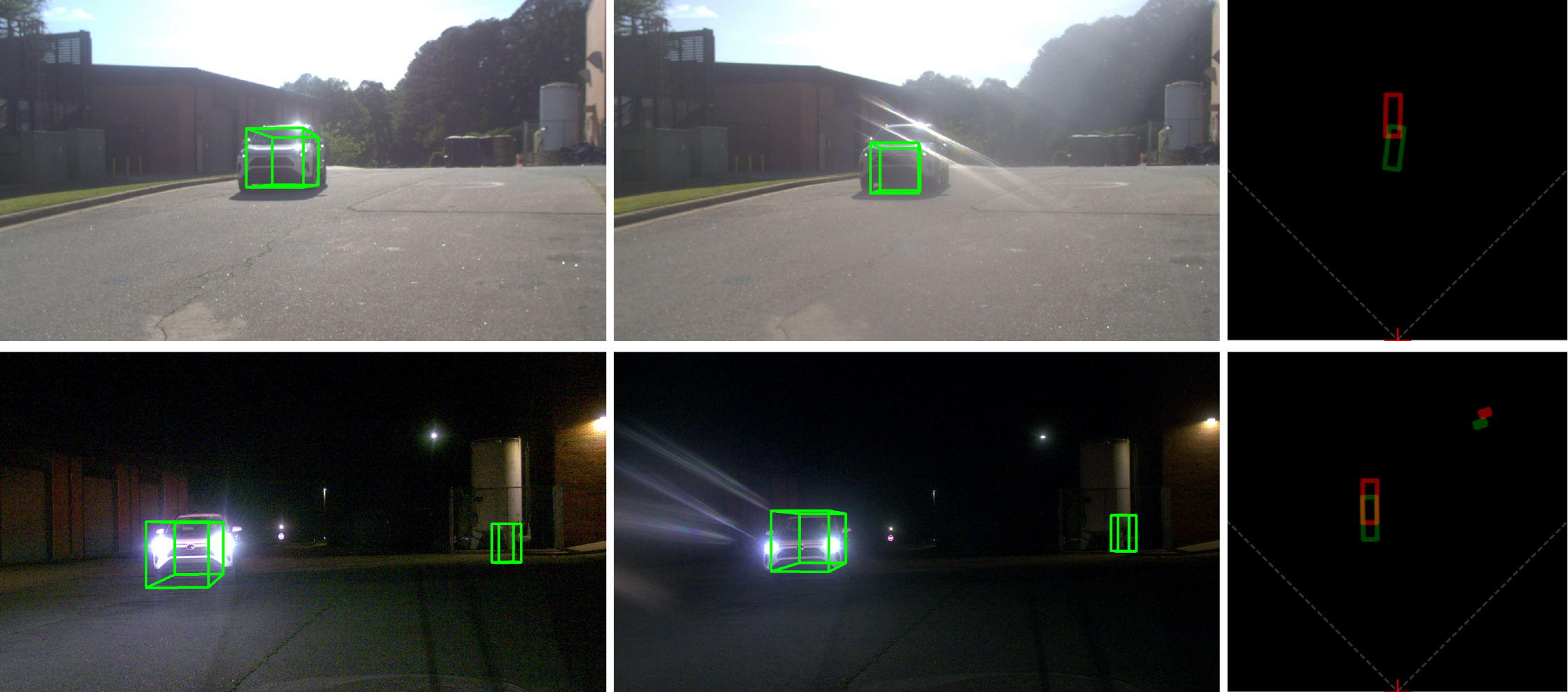}
    \caption{Representative OVM3D physical examples from farther-direction day and night scenes. Each row shows the clean prediction overlay, the attacked prediction overlay, and the BEV comparison.}
  \end{subfigure}
  \caption{Physical evaluation of an approaching vehicle under daytime and nighttime conditions. Both the monocular depth model and the 3D detector show clear depth shifts in the attacked views relative to the clean predictions, confirming that the scratch-induced streak artifact transfers from the digital renderer to real camera inputs and produces structured prediction changes across both tasks.}
  \label{fig:physical_qualitative}
  \vspace{-1em}
\end{figure}

\revision{%
The detailed trend also reveals that the effect is structured rather than random.
md4all exhibits a clear day/night asymmetry: daytime scenes produce a farther-oriented mid-range bias, whereas nighttime scenes produce a strong closer-oriented response over the same range.
OVM3D is more heterogeneous during the day, with both closer- and farther-oriented behaviors appearing across trajectories, but the dominant attack curves still remain above the clean fluctuation baseline in the effective range.
At night, the dominant OVM3D behavior is again a farther-oriented mid-range response, in some cases accompanied by substantial target loss.%
}{%
The detailed trend also reveals that the effect is structured rather than random.
For md4all during the day, the deployed scratch shifts the prediction farther in the mid range.
At night, it shifts the prediction closer to the same range, and by a larger amount.
In both conditions, the two materials trace the same curve, with glass reaching the larger magnitude at night.
For OVM3D during the day, we deploy scratches optimized in each direction, and each one moves the prediction the way it was optimized to, with either material.
At night, OVM3D shifts the prediction farther in the mid range, in some cases accompanied by substantial target loss, and the glass sessions separate from the baseline band more clearly than the acrylic ones.
Across all four panels, the two materials follow broadly the same trend, so the material mainly affects the magnitude of the shift, while its direction and the depth range where it is strongest stay largely the same.%
}
The farther-direction curves for OVM3D show a localized dip near 27\,m before recovering; this is likely a model-internal effect, as OVM3D couples depth prediction to 3D geometric consistency via a disentangled loss and per-category size priors, and the interplay between these geometric constraints and direct depth regression may produce less consistent predictions at the projected box scale corresponding to this distance. Figure~\ref{fig:physical_qualitative} shows representative physical examples from this effective range.

\revision{}{%
Two controls constrain the interpretation of these curves.
The dual-clean reference stays within about $1.5\%$ of zero during the day, while at night it sits at $-3.3\%$ for md4all and $-5.7\%$ for OVM3D.
Both nighttime values are negative, and they likely arise because the two cameras view the target's headlights from slightly different angles, so the intrinsic flare in each imaging path differs slightly and produces a steady systematic offset in the measurement.
Our physical experiments cannot isolate or remove this offset.
However, both models still show a substantial depth shift relative to the baseline band.
The no-trigger control further confirms that the shifts arise from the trigger rather than the scratched cover alone.
With the same cover mounted but the target's lights switched off, the mean relative change over the whole approach is $+0.1\%$ for md4all and $-1.6\%$ for OVM3D, and neither camera loses the target in any frame.
Both values are far below the shifts the same cover produces under an active trigger.
Accordingly, in the two nighttime panels of Figure~\ref{fig:phys_forward}, the no-trigger curves remain close to zero, while the attacked curves deviate substantially.
Thus, the scratched cover has a negligible optical effect without a compatible bright light source, keeping the attack dormant during normal operation.%
}

\revision{}{%
Because \proposed and the three most closely related physical attacks differ in attack surface, physical setup, victim model, and aggregation procedure, their reported results provide an order-of-magnitude comparison rather than a controlled head-to-head benchmark.
Specifically, Pi-Jack reports a depth shift of approximately 10.1\,m~\cite{299619_pi_jack}, 3D$^2$Fool reports 10.4\,m~\cite{zheng2024_3d2fool_physical_mde}, and LensAttack reports 2.7\,m~\cite{zhou2024optical} in their respective physical experiments.
Under our physical setup, \proposed causes shifts of approximately 4.7\,m during the day and 7.0\,m at night.
Thus, its physical effect falls within the same meter-level range as these prior attacks.
However, their attack mechanisms and deployment models differ.
Pi-Jack manipulates geometric cues through objects arranged in the observed scene, 3D$^2$Fool applies adversarial camouflage textures to target vehicles, and LensAttack uses a purpose-built optical attachment that continuously alters the camera input.
Accordingly, these approaches require either a visible modification of the observed scene or an additional optical element in front of the camera.
By comparison, \proposed uses a small passive scratch on the camera-side optical front-end, requires no attacker presence at runtime, and affects the image only when a compatible light source triggers the scratch-induced streak.
Therefore, \proposed combines a meter-level physical effect with a passive, trigger-dependent deployment model.%
}

Overall, the physical results show that the scratch-induced optical artifact transfers from the digital renderer to real camera inputs as a structured prediction change.

\revision{}{%
A separate physical experiment demonstrates that the attack extends beyond monocular perception.
As described in Appendix~\ref{sec:appendix_stereo}, we apply the scratch to only one of the two cameras in a commercial stereo camera, leaving the other camera clean.
Despite the clean second view, the attack remains effective.
In the region affected by the scratch-induced streak, the stereo system reports a continuous surface approximately 2\,m in front of the camera, even though the true surface is 21\,m away.
This result shows that the additional view provided by the stereo camera does not prevent the attack.%
}

% \by{update this title. what you want to convey here?}
% \ql{RESOLVED}

\subsection{\proposed Attack against Countermeasures}
\label{sec:countermeasures}
\revision{%
The attack surface exposed by SLASH raises the question of how camera-based systems can be defended.
We discuss two candidate defense directions and their practical limitations; detailed results are reported in Appendix~\ref{sec:appendix_countermeasures}.%
}{%
The attack surface exposed by SLASH raises the question of how camera-based systems can be defended.
We evaluate three candidate defense directions, reported in full in Appendix~\ref{sec:appendix_countermeasures}.
Two of the defenses target the attack through flare removal and through adversarial training.
The remaining one considers generic image enhancements that an autonomous driving system might use.%
}

\revision{%
Taken together, neither standard image-space nor training-time defense neutralizes \proposed by itself in our evaluation: flare removal risks suppressing legitimate scene content, while adversarial training can be circumvented by re-optimizing against the hardened model.%
}{%
None of the three neutralizes \proposed in our evaluation: flare removal reduces the depth shift but risks suppressing legitimate scene content, generic enhancement retains an attack effect and can amplify it, and adversarial training is bypassed by re-optimizing against the hardened model.%
}
\revision{%
A principled defense against passive optical tampering likely requires dedicated optical-path integrity primitives, such as multi-camera or temporal consistency checks, which we leave to future work.%
}{%
An effective defense against passive optical tampering likely needs to inspect the optical path itself, rather than the depth output it produces, thereby demonstrating the serious threat posed by \proposed.%
}

\section{Discussion and Limitations}
\label{sec:limitations}

\noindent\textbf{Model coverage.}
Our evaluation covers three monocular depth estimators and three monocular 3D detectors, and shows that detector-style architectures absorb the streak artifact noticeably better than dense depth regressors, likely because object-centric features and category size priors damp local image perturbations.
\revision{%
Each attack is also optimized against the same model that is evaluated; the low parameter dimensionality of $\theta$ and the geometric, rather than texture-based, nature of the streak make a deployed scratch unlikely to overfit a single model, but a quantitative cross-architecture transferability study is left to future work.%
}{%
Each attack is also optimized against the model it is evaluated on. Appendix~\ref{sec:appendix_transfer} measures transfer without re-optimization: the effect carries across depth models, but on 3D detectors it weakens into target loss instead of a controlled distance bias.%
}

\noindent\textbf{Geometric-optics renderer.}
We use a geometric-optics renderer rather than a full wave-optical simulator, which keeps the attack independent of lens-stack and ISP details that an attacker generally cannot identify, but does not capture diffraction-induced spectral dispersion at microstructured scratches. This contributes to a small residual gap between digital and physical $\Delta\mathrm{RE}$ magnitudes, which a learned or physics-based appearance prior could further close.

\noindent\textbf{Trigger requirement.}
\proposed activates only when a compatible compact bright source illuminates the scratched region, and is weak when the trigger is far from the projected scratch line.
This passive triggering is what makes the attack stealthy in normal operation, but it also concentrates the attack into scenes with headlights, specular highlights, or strong directional sunlight, which are precisely the conditions under which depth-aware perception is most safety-critical.
\revision{}{%
A scratch pattern is tied to the viewing geometry it was optimized for, which bounds the depth window and the range of viewing angles over which it is effective.
The scratch is persistent, so the target need only pass through that window on one of the repeated encounters that occur while the scratch remains in place.
As for other light sources in the environment, our physical experiments include light sources beyond the target's headlights that do not noticeably affect the attack, while our digital experiments annotate the light sources present in each scene and render their streaks.
However, the attack fails once enough light sources produce streaks of their own to cover the target region and mask the optimized streak pattern.%
}

\revision{}{%
\noindent\textbf{Camera and mounting assumptions.}
The scratch-to-streak mapping depends on camera front-end parameters that an attacker may know only approximately, since the threat model does not assume access to calibrate the deployed unit.
We assess this uncertainty by changing one parameter at a time from our physical camera setup (8.5\,mm/F2.8 lens, 3\,m focus, and 13\,mm distance between scratch and pupil).
A $3\%$ focal-length error changes the predicted streak length by only $6.1\%$, while refocusing from 3\,m to infinity changes it by $0.2\%$.
In contrast, increasing the distance between scratch and pupil from 13 to 15\,mm shortens the predicted streak by $13.4\%$, and increasing it to 20\,mm shortens the streak by $35.2\%$.
Thus, the tested focal-length and focus errors have little effect on the predicted streak length, whereas the distance between scratch and pupil is the main geometric sensitivity.
This distance is determined by the front-end geometry and can often be estimated externally for cameras with an exposed lens, protective cover, or similar front optical element.
Our physical experiments validate this setting by engraving scratches on covers mounted directly in front of the camera.
We have not tested windshield-mounted cameras, where curvature, coatings, camera-to-glass distance, and vehicle-specific mounting geometry may change the streak formation.
Therefore, our results support exposed or cover-mounted front-end attacks, while windshield-level validation remains future work.%
}

\revision{}{%
\noindent\textbf{Defenses beyond the image space.}
Our countermeasure study focuses on image-space defenses. In a commercial autonomous driving system, camera outputs may be fused with LiDAR or radar. Because \proposed corrupts only the camera optical path, its impact on a fused estimate depends on the fusion architecture and on how much weight the system assigns to the corrupted camera evidence. Visual redundancy within the camera modality may also help, but it does not eliminate the risk in all cases: in our stereo case study, scratching one camera still corrupts the resulting stereo depth output, as shown in Appendix~\ref{sec:appendix_stereo}. These observations suggest that stronger defenses should move beyond post-hoc image processing and explicitly assess whether the visual evidence is consistent with a healthy sensing front-end and with other available sensor signals. We leave the design and evaluation of such system-level defenses to future work.
}

\vspace{-0.5em}
\section{Conclusion}

We presented \proposed, a passive camera-side optical attack in which a scratch-like defect on the lens or protective cover acts as a fixed, trigger-conditioned perturbation to the image-formation process. Unlike scene-side patches, \proposed does not modify the observed object or environment. Unlike active optical injection, it requires no attacker action after deployment. Its effect appears when compact scene light sources interact with the scratched front-end and produce localized streak artifacts that corrupt depth-relevant image cues.

\proposed formulates this threat as an optical-space optimization problem with three components: a geometry-constrained renderer that maps scratch parameters and trigger locations to streak center, orientation, and length; an appearance synthesizer that renders the resulting light-like artifact; and a scene-specific optimizer that searches for a scratch pattern that remains effective across a sequence of frames. Digital experiments show that scene-specific optimization can produce substantial directional target-depth shifts on monocular depth estimation and monocular 3D object detection. Physical experiments with paired clean and scratched camera views further show that the effect transfers to real camera inputs as structured prediction changes under compatible illumination.

These findings highlight optical-path integrity as a security assumption for camera-based perception. A small defect on the accessible camera front-end can behave as a latent, scene-triggered adversarial mechanism even when the scene, model, and software pipeline are unchanged.

\section*{Acknowledgment}
This work was supported in part by the National Science Foundation under Grant Nos. 2427316 and 2444389.

\bibliographystyle{IEEEtran}
\bibliography{bib/adv.bib,bib/monodepth.bib,bib/3ddet.bib,bib/other.bib,bib/defense.bib}

\appendices
% \appendices leaves \thesection empty, so IEEEtran's default subsection label
% (\thesection-\Alph{subsection}) renders cross-references as "-A". Match the
% printed heading, which is "A.".
\renewcommand{\thesubsection}{\mbox{\Alph{subsection}}}
\renewcommand{\thesubsubsection}{\thesubsection\arabic{subsubsection}}

% Deleted section, rendered only in the diff view. A sectioning command and a
% float cannot appear inside \revision's argument, so the deletion is guarded
% by \ifrevisionshowdiff instead.
\suppressfloats[t]
\ifrevisionshowdiff
{
\color{revdelcolor}

\subsection{Anisotropic Scattering: Physical Intuition}
\label{sec:appendix_aniso}

Figure~\ref{fig:aniso_btdf} shows the angular distribution of transmitted light through a microfacet surface, illustrating why a scratch-induced artifact takes the form of a directional streak rather than an isotropic blur.
A scratched or grooved surface is not uniformly rough in all directions: its dominant roughness axis causes transmitted light to spread preferentially along that axis, producing an elongated lobe rather than a symmetric halo.
This anisotropic behavior motivates modeling the artifact as a streak aligned with the projected scratch axis \(\hat{\mathbf{s}}_k\), as described in \S\ref{subsec:geometry_mapping}.
The renderer does not use the GGX model directly; the figure serves as physical intuition only.
}

\begin{figure}[!t]
\centering
\includegraphics[width=0.9\linewidth]{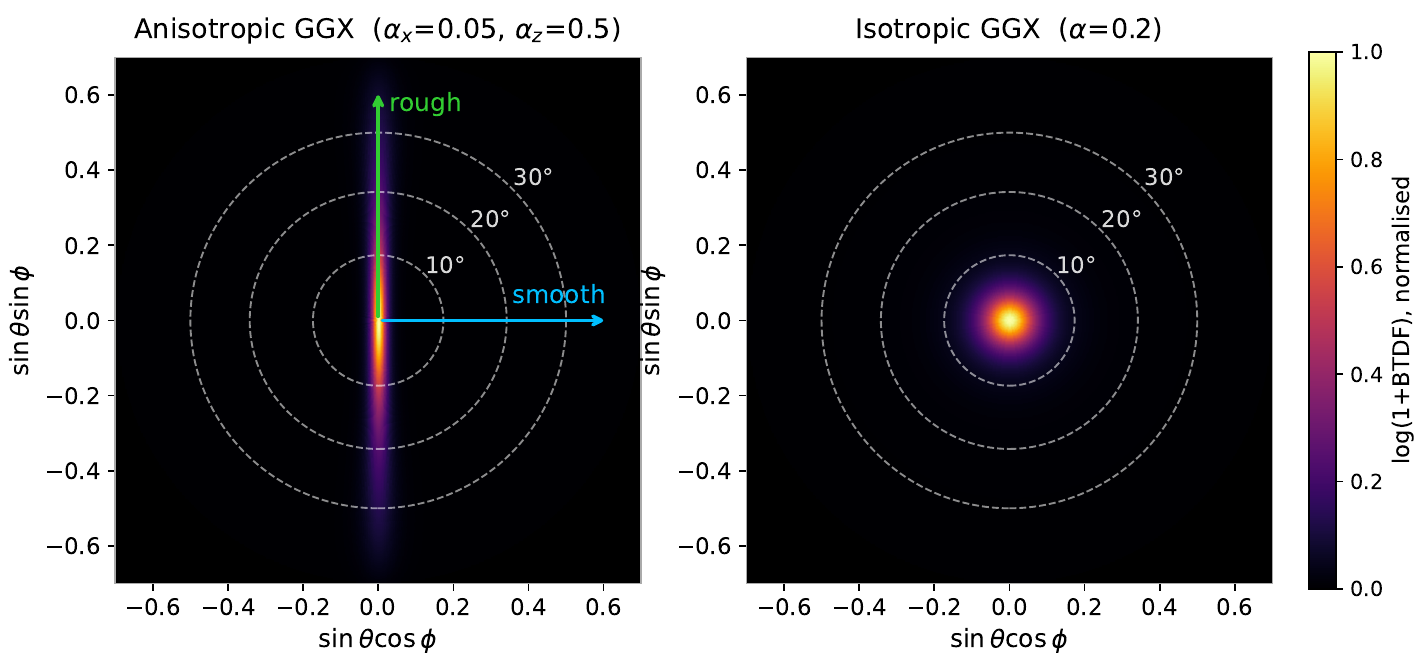}
\caption{{\color{revdelcolor}\textbf{Anisotropic scattering intuition.} The panels show the angular distribution of transmitted light through a microfacet surface. Left: an anisotropic GGX~\cite{walter2007microfacet} surface with different roughness values along two axes produces a directional lobe. Right: an isotropic GGX surface produces a more symmetric lobe. This supports modeling a scratch-induced artifact as a directional streak rather than an isotropic blur.}}
\label{fig:aniso_btdf}
\end{figure}
\fi

% \by{the appendix text shuold be at the top of the left column in this page. }
% \ql{RESOLVED}
\begin{center}
\textsc{Appendix}
\end{center}
% \by{do not use 'section', use 'subsection'}
% \ql{RESOLVED}
\ifrevisionshowdiff
{
\color{revdelcolor}
\subsection{Additional 3D Detection Results}
\label{sec:appendix_det3d_results}

This subsection provides qualitative examples and single-frame optimization results for 3D object detection on the nuScenes dataset.

\ifrevisionshowdiff
{
\color{revdelcolor}
\noindent\textbf{Scene-specific OVM3D qualitative examples.}
Figure~\ref{fig:ovm3d_per_sequence_qualitative} shows representative OVM3D~\cite{huang2024_ovm3d} attacks in which the target vehicle remains detected but shifts in depth.
}

\begin{figure}[t]
  \centering
  \includegraphics[width=0.85\columnwidth]{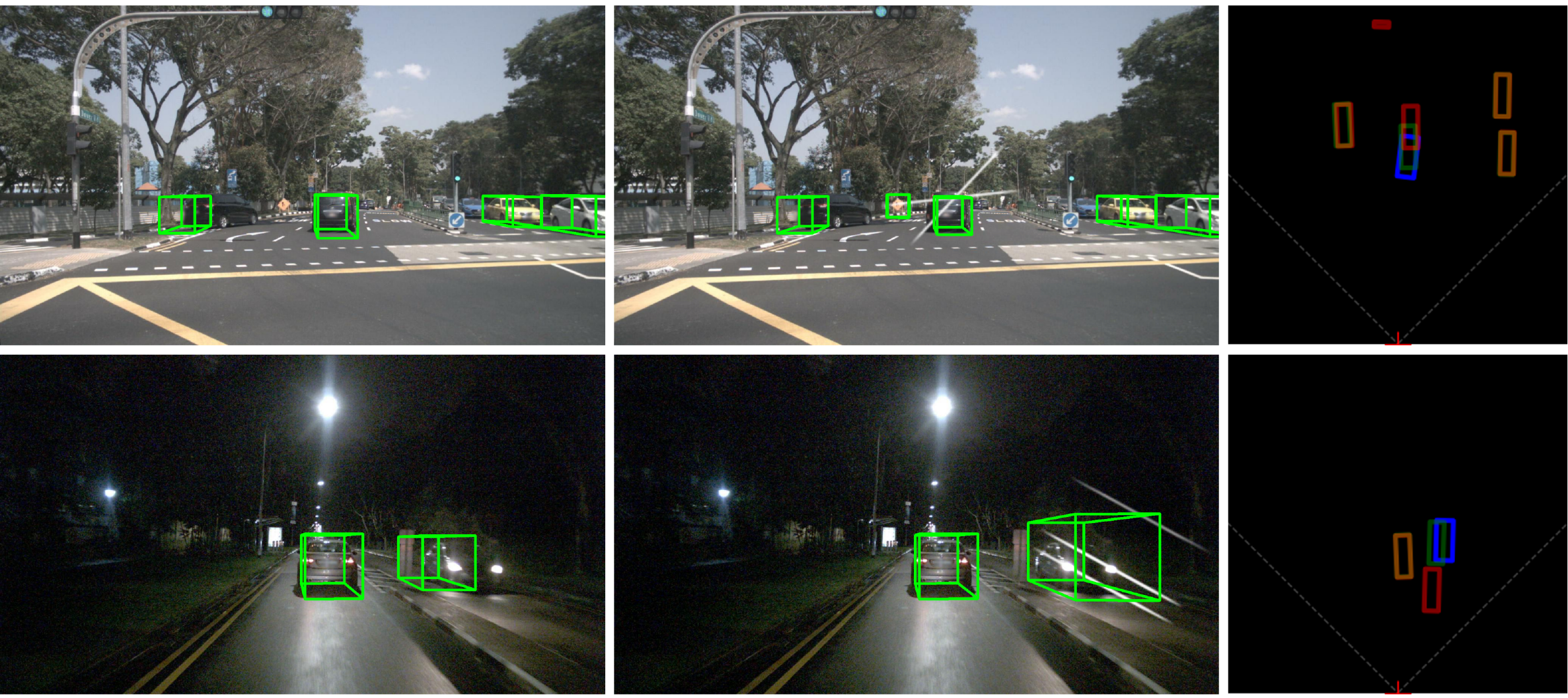}
  \caption{{\color{revdelcolor}Qualitative examples of attack-induced depth shifts in 3D object detection. Each row shows an OVM3D~\cite{huang2024_ovm3d} example in which the target remains detected. The columns show the clean prediction overlay, attacked prediction overlay, and bird's-eye-view comparison. Clean predictions are green, attacked predictions are red, ground truth is blue, and unchanged boxes are orange.}}
  \label{fig:ovm3d_per_sequence_qualitative}\vspace{-1em}
\end{figure}
\fi

\noindent\textbf{Single-frame optimization.}
Table~\ref{tab:det3d_pervehicle} reports the single-frame evaluation on nuScenes, in which scratch parameters are optimized independently for each frame.

\begin{table}[b]
  \centering
  \caption{Single-frame attack results on 3D object detection. Scratch parameters are optimized independently per frame, so this setting is a non-deployable upper bound. C and F denote closer and farther attack directions, and TLR is defined as in Table~\ref{tab:det3d_persequence}.}
  \label{tab:det3d_pervehicle}
  \small
  \resizebox{\columnwidth}{!}{%
  \begin{tabular}{ll S[table-format=+1.2] S[table-format=+2.2] S[table-format=+2.2] S[table-format=2.2]}
    \toprule
    \multirow{2}{*}{Model} & \multirow{2}{*}{Scene} & \multicolumn{1}{c}{\multirow{2}{*}{Clean RE}} & \multicolumn{2}{c}{$\Delta$RE} & \multicolumn{1}{c}{\multirow{2}{*}{TLR}} \\
    \cmidrule(lr){4-5}
    & & & \multicolumn{1}{c}{Closer} & \multicolumn{1}{c}{Farther} & \\
    \midrule
    \multirow{2}{*}{FCOS3D}
      & Day   & +2.70 & -3.59 & +6.38 & 7.75 \\
      & Night & -0.42 & -4.42 & +13.22 & 1.04 \\
    \midrule
    \multirow{2}{*}{PGD}
      & Day   & +2.14 & -3.45 & +5.62 & 5.54 \\
      & Night & -0.74 & -5.11 & +12.15 & 1.40 \\
    \midrule
    \multirow{2}{*}{OVM3D}
      & Day   & -4.06 & -7.89 & +6.48 & 6.51 \\
      & Night & -6.56 & -15.64 & +15.77 & 1.27 \\
    \bottomrule
  \end{tabular}%
  }
\end{table}
}
\fi

\subsection{Streak Appearance Synthesis}
\label{sec:appendix_appearance}

This subsection derives the defocus-limited streak length and specifies the renderer.
For object distance \(d\), the signed Gaussian image distance is \(z(d)=fd/(d-f)\); the sensor is at \(z_f=z(d_f)\), and the scratched front-end has image distance \(z_s=z(d_s)\).
With aperture diameter \(A=f/N_f\) and f-number \(N_f\), the circle-of-confusion diameter and its pixel length are
\begin{gather}
c_{\mathrm{sensor}} = A\frac{|z_f-z_s|}{|z_s|},
\label{eq:coc}\\
\ell_{\mathrm{px}} = \frac{c_{\mathrm{sensor}}}{\rho}, \qquad \rho = \frac{1}{2}\left(\frac{F_w}{W}+\frac{F_h}{H}\right),
\label{eq:streak-length-px}
\end{gather}
where \(\rho\) is the average pixel pitch.
The defocus-limited \(\ell_{\mathrm{px}}\) bounds the streak length when the scratch is much closer to the lens than the focused scene.

For each trigger-scratch pair \((m,k)\), the geometric mapping gives streak center \(c_{mk}\) and direction \(\hat{\mathbf{s}}_k\).
The renderer rasterizes a rectangular mask \(R_{mk}\) of length \(\ell_{\mathrm{px}}\) and normalized width \(w\), centered at \(c_{mk}\) and aligned with \(\hat{\mathbf{s}}_k\), then softens it with isotropic and directional blurs:
\begin{equation}
\widetilde{R}_{mk} = R_{mk} * G_\sigma * B_{\hat{\mathbf{s}}_k,b_{\mathrm{blur}}},
\label{eq:per-pair-mask}
\end{equation}
where \(G_\sigma\) is a Gaussian kernel with standard deviation \(\sigma\), \(B_{\hat{\mathbf{s}}_k,b_{\mathrm{blur}}}\) is a one-dimensional blur kernel of extent \(b_{\mathrm{blur}}\) along \(\hat{\mathbf{s}}_k\), and \(*\) denotes 2D convolution.

A trigger farther from the projected scratch line should produce a weaker streak because less of its light interacts with the scratched front-end.
Let \(\hat{\mathbf{n}}_k=(-\hat{s}_{k,y},\hat{s}_{k,x})\) be perpendicular to the projected scratch direction.
The distance from trigger \(m\) to scratch \(k\) is
\begin{equation}
d_{mk}=\left|(r_m-q_k)^\top\hat{\mathbf{n}}_k\right|.
\end{equation}
We model the attenuation as
\begin{equation}
\eta_{mk} = \exp\left(-\frac{d_{mk}^2}{2\sigma_d^2}\right).
\label{eq:streak-attenuation}
\end{equation}
The attenuated mask is \(M_{mk}=\eta_{mk}\widetilde{R}_{mk}\).

We combine the trigger-scratch contributions using screen blending as a bounded approximation of additive brightening rather than an opaque overlay:
\begin{gather}
\mathcal{M} = 1-\prod_{m=1}^{|L|}\prod_{k=1}^{K}(1-M_{mk}),
\label{eq:mask-composite}\\
\begin{gathered}
\overline{\mathcal{M}} = \mathrm{clip}(s\,\mathcal{M},0,1),\\
I' = 1-(1-I)\otimes(1-\overline{\mathcal{M}}),
\end{gathered}
\label{eq:image-composite}
\end{gather}
where \(s\) is a global brightness scale and \(\otimes\) denotes element-wise multiplication.
The appearance parameters are \(\{w,\sigma,b_{\mathrm{blur}},\sigma_d,s\}\).

\subsection{Implementation and Annotation Details}
\label{sec:appendix_impl}

This subsection provides the annotation procedure, renderer settings, and search space used in experiments on the nuScenes dataset.

\noindent\textbf{Trigger annotation.}
For each frame, we manually annotate compact bright regions on the target vehicle as trigger locations for the scratch image-formation model.
The main light sources are the target's headlights at night and specular reflections from the body or windshield during the day.
We pair these annotations with the target's nuScenes 3D bounding box, which provides the target region used by the attack objective and the LiDAR-derived depth used for evaluation.

\noindent\textbf{Renderer and search space.}
The renderer uses the parameters of the Basler acA1920-48gc camera and Edmund Optics 8.5\,mm/F2.8 lens used in our physical experiments.
Each attack uses $K=2$ scratches and jointly optimizes their $(\alpha_k,b_k)$ parameters, giving a 4-dimensional search space.
The scratch angle $\alpha_k$ spans $[0,360^{\circ})$, and the offset $b_k$ ranges symmetrically around the optical axis.

\subsection{Model-Specific Evaluation Details}
\label{sec:appendix_model_eval}

This subsection provides the task-specific evaluation and depth-calibration procedures used for monocular depth estimation and 3D object detection on the nuScenes dataset.

\noindent\textbf{3D object detection.}
For FCOS3D and PGD, both the front and back cameras are used. The two per-camera sample sets are concatenated and the metric is macro-averaged across the combined set rather than averaging two predictions per frame; FCOS3D and PGD are trained on all nuScenes cameras and exhibit consistent accuracy across both. OVM3D results follow the same pooling convention.

\noindent\textbf{3D detection target identification.}
\label{sec:appendix_det3d_match}
Scoring an attack outcome requires identifying whether the intended target remains detected after the attack.
We first remove predictions whose projected 2D boxes overlap strongly with non-target ground-truth boxes.
Among the remaining predictions, we select the one closest to the target ground-truth center subject to sufficient 2D IoU with the target's projected box, and reject it if its 3D IoU with a non-target ground-truth box is too high.
When this procedure returns no candidate, the sample is treated as a target loss and contributes to TLR rather than to the depth-shift metric.

\noindent\textbf{Monocular depth estimation.}
All three monocular depth estimators are evaluated on the front camera, consistent with the cameras used to train MonoDepth2~\cite{godard2019_monodepth2} and md4all~\cite{Gasperini_2023_ICCV_md4all}.
MonoDepth2 outputs disparity, which we convert to metric depth using the standard KITTI~\cite{geiger2013_kitti} stereo calibration scale.
The md4all model is supervised with metric depth labels, so we use its output without additional scaling.
DCPI-Depth~\cite{11051129_dcpi_depth} is scale-ambiguous.
For target-centric RE, we expand the target bounding box by a factor of two, collect LiDAR-projected pixels within this region, and compute a local scale factor from the median LiDAR depth and median clean predicted depth.
We apply the same clean-derived scale to both the clean and attacked predictions before computing RE.
This procedure calibrates the prediction to the target region without using information from the attacked prediction.

\subsection{Optimizer Configuration and Comparison}
\label{sec:appendix_optimizer}

This subsection provides the optimizer configuration and evaluates how optimizer choice, population size, and evaluation budget affect the attack on the nuScenes dataset.

\revision{%
The main experiments use CMA-ES~\cite{hansen2016cma} with a population of $20$, an initial step size of $\sigma_0=0.15$ in the normalized $[0,1]^4$ search space, and a budget of $100$ objective evaluations per instance.%
}{%
In Table~\ref{tab:appendix_optimizer_summary}, Ours denotes the default CMA-ES~\cite{hansen2016cma} setting used in the main experiments: a population of $20$, five generations, an initial step size of $\sigma_0=0.15$ in the normalized $[0,1]^4$ search space, and $100$ objective evaluations per instance.%
}

\revision{%
We tested several popular alternatives and found that population-based optimizers, including CMA-ES, Differential Evolution~\cite{storn1997differential}, and Particle Swarm Optimization~\cite{kennedy1995particle}, produced comparable attack effects, while L-BFGS~\cite{liu1989limited} was consistently weaker. We therefore use CMA-ES as the default because it is stable, gradient-free, and effective within a small evaluation budget.%
}{%
Table~\ref{tab:appendix_optimizer_summary} compares CMA-ES with Differential Evolution (DE)~\cite{storn1997differential}, Particle Swarm Optimization (PSO)~\cite{kennedy1995particle}, and Limited-memory Broyden-Fletcher-Goldfarb-Shanno (L-BFGS)~\cite{liu1989limited} and reports representative population-size and evaluation-budget results for MonoDepth2.
CMA-ES, DE, and PSO produce comparable attack effects, but CMA-ES uses $100$ objective evaluations per sequence versus $360$ for DE and $440$ for PSO; L-BFGS is weaker despite $450$ evaluations.
Therefore, we use CMA-ES to balance optimization cost and attack effectiveness.%
}

\ifrevisionshowdiff
\begin{table}[t]
  \centering
  \color{revdelcolor}
  \caption{{\color{revdelcolor}Attack effectiveness across optimization methods. All methods are evaluated with MonoDepth2 on the same front-camera target sequences. $\Delta\mathrm{RE}$ is reported as a percentage, and the last row reports the objective-evaluation budget per sequence.}}
  \label{tab:appendix_optimizer}
  \small
  \begin{tabular}{cc S[table-format=+2.2] S[table-format=+2.2] S[table-format=+2.2] S[table-format=+2.2]}
    \toprule
    Scene & Direction & \multicolumn{1}{c}{CMA-ES} & \multicolumn{1}{c}{DE} & \multicolumn{1}{c}{PSO} & \multicolumn{1}{c}{L-BFGS} \\
    \midrule
    \multirow{2}{*}{Day}
      & Closer  & -20.34 & -22.15 & -20.96 & -16.75 \\
      & Farther & +14.32 & +17.28 & +17.07 & +6.63 \\
    \midrule
    \multirow{2}{*}{Night}
      & Closer  & -29.12 & -33.29 & -32.98 & -28.56 \\
      & Farther & +8.52 & +10.82 & +11.10 & +5.18 \\
    \midrule
    \multicolumn{2}{l}{Evaluations} & \multicolumn{1}{c}{100} & \multicolumn{1}{c}{360} & \multicolumn{1}{c}{440} & \multicolumn{1}{c}{450} \\
    \bottomrule
  \end{tabular}
\end{table}
\fi

\ifrevisionshowdiff
{
\color{revdelcolor}
\subsubsection*{Population Size and Evaluation Budget}

We conduct separate sweeps of population size and evaluation budget.
Each sweep holds the other parameter fixed.
Both sweeps use MonoDepth2 with the front camera under scene-specific optimization.

Table~\ref{tab:appendix_popsize} reports the population-size sweep from $10$ to $50$ at the same total cost.
Across this fivefold change, three of the four conditions differ by at most $1.7$ percentage points, while the nighttime closer result differs by $3.6$ percentage points.
The attack effect is therefore consistent across the tested population sizes.

Table~\ref{tab:appendix_budget} reports the evaluation-budget sweep at a fixed population size of $20$.
Increasing the budget from $100$ to $200$ changes the attack effect by an average of $2.7$ percentage points across the four conditions, with the nighttime closer result changing from $-29.12\%$ to $-33.39\%$.
Thus, the attack effect remains broadly consistent across the tested budgets, while a larger offline budget can further strengthen the attack.
}
\fi

\revision{}{%
We sweep population size at a fixed budget of $100$ evaluations and evaluation budget at a fixed population of $20$.
Population-size settings in Table~\ref{tab:appendix_optimizer_summary} are (population, generations) pairs.
Across populations from $10$ to $50$, three of the four conditions differ by at most $1.7$ percentage points, while the nighttime closer result differs by $3.6$ percentage points.
Increasing the budget from $100$ to $200$ changes the attack effect by an average of $2.7$ percentage points across the four conditions, with the nighttime closer result changing from $-29.12\%$ to $-33.39\%$.
These results show that the attack effect persists across the tested settings, while a larger offline budget can strengthen it.%
}

\ifrevisionshowdiff
\begin{table}[t]
  \centering
  \color{revdelcolor}
  \caption{{\color{revdelcolor}Sensitivity to CMA-ES population size. The sweep evaluates MonoDepth2 on front-camera target sequences with a fixed budget of $100$ objective evaluations. Each column is a (population, generations) pair with the same total cost. $\Delta\mathrm{RE}$ is reported as a percentage.}}
  \label{tab:appendix_popsize}
  \small
  \begin{tabular}{cc S[table-format=+2.2] S[table-format=+2.2] S[table-format=+2.2] S[table-format=+2.2]}
    \toprule
    Scene & Direction & \multicolumn{1}{c}{$(10, 10)$} & \multicolumn{1}{c}{$(20, 5)^\dagger$} & \multicolumn{1}{c}{$(25, 4)$} & \multicolumn{1}{c}{$(50, 2)$} \\
    \midrule
    \multirow{2}{*}{Day}
      & Closer  & -20.91 & -20.34 & -19.97 & -20.02 \\
      & Farther & +15.46 & +14.32 & +13.72 & +14.66 \\
    \midrule
    \multirow{2}{*}{Night}
      & Closer  & -32.72 & -29.12 & -29.61 & -31.22 \\
      & Farther & +9.66  & +8.52  & +8.78  & +10.11 \\
    \bottomrule
  \end{tabular}
  \\[0.3em]
  \footnotesize $\dagger$ default setting used in all main experiments.
\end{table}
\fi

\ifrevisionshowdiff
\begin{table}[t]
  \centering
  \color{revdelcolor}
  \caption{{\color{revdelcolor}Sensitivity to the CMA-ES evaluation budget. The sweep evaluates MonoDepth2 on front-camera target sequences with a fixed population of $20$. Columns report the number of objective evaluations per sequence. $\Delta\mathrm{RE}$ is reported as a percentage.}}
  \label{tab:appendix_budget}
  \small
  \setlength{\tabcolsep}{4pt}
  \resizebox{\columnwidth}{!}{%
  \begin{tabular}{cc *{6}{S[table-format=+2.2]}}
    \toprule
    Scene & Direction & \multicolumn{1}{c}{$20$} & \multicolumn{1}{c}{$40$} & \multicolumn{1}{c}{$60$} & \multicolumn{1}{c}{$80$} & \multicolumn{1}{c}{$100^\dagger$} & \multicolumn{1}{c}{$200$} \\
    \midrule
    \multirow{2}{*}{Day}
      & Closer  & -19.29 & -18.06 & -19.40 & -20.64 & -20.34 & -21.99 \\
      & Farther & +9.48  & +11.71 & +13.48 & +13.08 & +14.32 & +16.44 \\
    \midrule
    \multirow{2}{*}{Night}
      & Closer  & -23.90 & -27.58 & -30.29 & -29.72 & -29.12 & -33.39 \\
      & Farther & +5.65  & +6.54  & +9.35  & +8.82  & +8.52  & +11.29 \\
    \bottomrule
  \end{tabular}
  }
  \\[0.3em]
  \footnotesize $\dagger$ default setting used in all main experiments.
\end{table}
\fi

\begin{table}[t]
  \centering
  \ifrevisionshowdiff\color{revaddcolor}\fi
  \caption{\revision{}{Effects of optimizer choice, population size, and evaluation budget on attack effectiveness. Values are $\Delta\mathrm{RE}$ percentages for MonoDepth2 on front-camera target sequences. Day/Night are lighting conditions, and Closer/Farther are attack directions. Ours is the default CMA-ES setting (population $20$, $5$ generations, $100$ evaluations).}}
  \label{tab:appendix_optimizer_summary}
  \footnotesize
  \setlength{\tabcolsep}{3pt}
  \begin{tabular}{@{}ll *{4}{S[table-format=+2.2]}@{}}
    \toprule
    \multirow[c]{2}{*}{Control} & \multirow[c]{2}{*}{Setting} & \multicolumn{2}{c}{Day} & \multicolumn{2}{c}{Night} \\
    \cmidrule(lr){3-4}\cmidrule(lr){5-6}
    & & \multicolumn{1}{c}{Closer} & \multicolumn{1}{c}{Farther} & \multicolumn{1}{c}{Closer} & \multicolumn{1}{c}{Farther} \\
    \midrule
    \multirow{3}{*}{Optimizer}
      & DE     & -22.15 & +17.28 & -33.29 & +10.82 \\
      & PSO    & -20.96 & +17.07 & -32.98 & +11.10 \\
      & L-BFGS & -16.75 & +6.63  & -28.56 & +5.18 \\
    \midrule
    \multirow{3}{*}{Population size}
      & $(10,10)$ & -20.91 & +15.46 & -32.72 & +9.66 \\
      & $(25,4)$  & -19.97 & +13.72 & -29.61 & +8.78 \\
      & $(50,2)$  & -20.02 & +14.66 & -31.22 & +10.11 \\
    \midrule
    \multirow{5}{*}{Evaluation budget}
      & 20  & -19.29 & +9.48  & -23.90 & +5.65 \\
      & 40  & -18.06 & +11.71 & -27.58 & +6.54 \\
      & 60  & -19.40 & +13.48 & -30.29 & +9.35 \\
      & 80  & -20.64 & +13.08 & -29.72 & +8.82 \\
      & 200 & -21.99 & +16.44 & -33.39 & +11.29 \\
    \midrule
    Ours & & -20.34 & +14.32 & -29.12 & +8.52 \\
    \bottomrule
  \end{tabular}
\end{table}

\ifrevisionshowdiff
{
\color{revdelcolor}
\subsection{Scratch Hyperparameter Ablation}
\label{sec:ablation_scratch}

This subsection evaluates how streak brightness and width affect attack effectiveness and image-level stealthiness on nuScenes.
We sweep each parameter separately with MonoDepth2 in the nighttime, farther-direction setting on the front camera, holding the other at its default value (brightness\,=\,1.0, width\,=\,0.0075).
Table~\ref{tab:ablation_scratch} reports mean $\Delta\mathrm{RE}$ and peak signal-to-noise ratio (PSNR) over the nighttime samples.
\revision{%
Figure~\ref{fig:scratch_param_ablation} shows the corresponding appearance on a representative nighttime scene with two headlights.%
}{}

Increasing brightness increases the depth shift while decreasing PSNR.
Increasing width also decreases PSNR, with the largest depth shift at width 0.015.
The default brightness and width retain a $+10.43\%$ depth shift at 28.50\,dB PSNR.

\begin{table}[t]
  \centering
  \caption{Effects of streak brightness and width on attack effectiveness and stealthiness. The evaluation uses MonoDepth2 in the nighttime farther direction. We sweep brightness and width separately. When one parameter is swept, the other remains at its default value ($\dagger$). $\Delta\mathrm{RE}$ is reported as a percentage. PSNR compares the attacked and clean images in dB, with higher values indicating less visible perturbation.}
  \label{tab:ablation_scratch}
  \small
  \begin{tabular}{l c S[table-format=+2.2] S[table-format=2.2]}
    \toprule
    Setting & Value & \multicolumn{1}{c}{$\Delta\mathrm{RE}$ $\uparrow$} & \multicolumn{1}{c}{PSNR (dB) $\uparrow$} \\
    \midrule
    \multirow{4}{*}{Brightness ($\times$)} &
      \tablenum[table-format=1.4]{0.5} & +10.14 & 35.40 \\
    & \tablenum[table-format=1.4]{1.0}\rlap{$^\dagger$} & +10.43 & 28.50 \\
    & \tablenum[table-format=1.4]{2.0} & +11.75 & 24.73 \\
    & \tablenum[table-format=1.4]{4.0} & +13.31 & 21.84 \\
    \midrule
    \multirow{4}{*}{Width} &
      \tablenum[table-format=1.4]{0.004} & +10.59 & 32.91 \\
    & \tablenum[table-format=1.4]{0.0075}\rlap{$^\dagger$} & +10.43 & 28.50 \\
    & \tablenum[table-format=1.4]{0.015} & +12.81 & 26.38 \\
    & \tablenum[table-format=1.4]{0.030} & +11.39 & 22.95 \\
    \bottomrule
  \end{tabular}
\end{table}

\ifrevisionshowdiff
\begin{figure*}[t]
  \centering
  \includegraphics[width=0.72\textwidth]{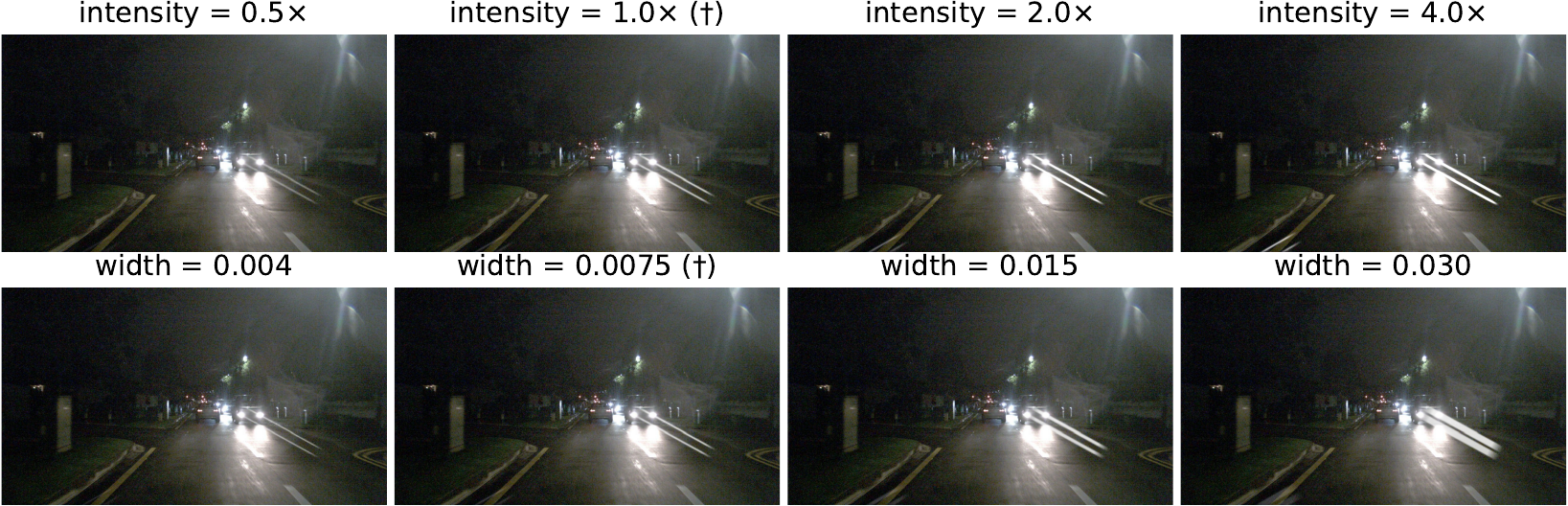}
  \caption{{\color{revdelcolor}Visual effects of streak brightness and width. The top row sweeps brightness, and the bottom row sweeps width on a representative nighttime scene with two vehicle headlights. Orange borders mark the default setting ($\dagger$).}}
  \label{fig:scratch_param_ablation}
\end{figure*}
\fi
}
\fi

\subsection{\revision{}{Number of Scratches}}
\label{sec:appendix_kscratch}

\revision{}{%
This subsection evaluates how the number of scratches affects attack effectiveness and stealthiness on the nuScenes dataset.
The sweep uses MonoDepth2 and the front camera under scene-specific optimization.
We sweep $K$ from $1$ to $4$ using the same budget of $100$ objective evaluations for every setting.
The $K=2$ row reproduces the main result.

Table~\ref{tab:appendix_kscratch_mde} shows that the $K=1$ single scratch setting already produces a substantial effect, reaching $-22.97\%$ in the nighttime closer direction.
Additional scratches improve attack effectiveness in most conditions, although the gains diminish at $K=4$.
Meanwhile, PSNR decreases monotonically in every condition as $K$ increases, showing that the streak artifacts become more visible.
We therefore use $K=2$ in the main experiments to balance attack effectiveness and stealthiness.%
}

\begin{table}[t]
  \centering
  \caption{\revision{}{%
Effects of scratch count on attack effectiveness and stealthiness. The sweep evaluates MonoDepth2 on front-camera target sequences. Every $K$ uses a budget of $100$ objective evaluations. $\Delta\mathrm{RE}$ is reported as a percentage, PSNR is reported in dB, and higher PSNR indicates greater stealthiness.%
}}
  \label{tab:appendix_kscratch_mde}
  \small
  \resizebox{\columnwidth}{!}{%
  \begin{tabular}{c S[table-format=+2.2] S[table-format=+2.2] S[table-format=2.2] S[table-format=2.2] S[table-format=+2.2] S[table-format=+2.2] S[table-format=2.2] S[table-format=2.2]}
    \toprule
    & \multicolumn{4}{c}{Day} & \multicolumn{4}{c}{Night} \\
    \cmidrule(lr){2-5}\cmidrule(lr){6-9}
    & \multicolumn{2}{c}{$\Delta$RE} & \multicolumn{2}{c}{PSNR}
    & \multicolumn{2}{c}{$\Delta$RE} & \multicolumn{2}{c}{PSNR} \\
    \cmidrule(lr){2-3}\cmidrule(lr){4-5}\cmidrule(lr){6-7}\cmidrule(lr){8-9}
    $K$ & \multicolumn{1}{c}{Closer} & \multicolumn{1}{c}{Farther} & \multicolumn{1}{c}{Closer} & \multicolumn{1}{c}{Farther} & \multicolumn{1}{c}{Closer} & \multicolumn{1}{c}{Farther} & \multicolumn{1}{c}{Closer} & \multicolumn{1}{c}{Farther} \\
    \midrule
    \tablenum[table-format=1.0]{1} & -15.17 & +12.70 & 34.73 & 36.43 & -22.97 & +8.47 & 30.04 & 32.56 \\
    \tablenum[table-format=1.0]{2}\rlap{$^\dagger$} & -20.34 & +14.32 & 31.73 & 33.92 & -29.12 & +8.52 & 27.00 & 29.44 \\
    \tablenum[table-format=1.0]{3} & -23.36 & +16.12 & 30.36 & 32.59 & -36.11 & +9.11 & 25.68 & 27.31 \\
    \tablenum[table-format=1.0]{4} & -25.72 & +15.63 & 29.19 & 31.32 & -40.30 & +8.81 & 24.31 & 25.92 \\
    \bottomrule
  \end{tabular}%
  }
  \\[0.3em]
  \footnotesize $\dagger$ default setting used in all main experiments.
\end{table}

\subsection{\revision{}{Sensitivity to Imperfect Placement}}
\label{sec:appendix_placement}

\revision{}{%
This subsection evaluates how differences between the optimized and engraved scratch placement affect the attack on the nuScenes dataset.
The evaluation uses MonoDepth2 and the front camera under scene-specific optimization.
Hand engraving can introduce error in the optimized scratch angle $\alpha_k$ and offset $b_k$.
We perturb each optimized scratch parameter by the stated magnitude with a random sign, render the resulting streaks, and run inference without re-optimization.
Each error is sampled once per sequence and remains unchanged across its frames, consistent with using the same scratch pattern throughout the target sequence.
The zero-error control reproduces the corresponding nuScenes results without placement error.

Table~\ref{tab:appendix_placement} shows that the attack tolerates moderate errors in both scratch angle and offset.
At a $5^\circ$ angle error, all four conditions retain at least $76\%$ of their zero-error effects.
At an offset error of $0.03$, all four conditions retain at least $60\%$.
Together, these results show that the attack retains substantial effects under moderate engraving imprecision.%
}

\begin{table}[t]
  \centering
  \caption{\revision{}{%
Sensitivity to scratch engraving error. Values are MonoDepth2 front-camera $\Delta\mathrm{RE}$ (\%); $\alpha$ is in degrees, $b$ uses normalized search-space units, and No error is the unperturbed control.%
}}
  \label{tab:appendix_placement}
  \footnotesize
  \setlength{\tabcolsep}{2pt}
  \ifrevisionshowdiff\color{revaddcolor}\fi
  \begin{tabular}{ll *{7}{S[table-format=+2.2]}}
    \toprule
    \multirow{2}{*}{Scene} & \multirow{2}{*}{Direction} & \multicolumn{1}{c}{\multirow{2}{*}{No error}} & \multicolumn{3}{c}{Angle error $\alpha$} & \multicolumn{3}{c}{Offset error $b$} \\
    \cmidrule(lr){4-6}\cmidrule(lr){7-9}
    & & & \multicolumn{1}{c}{$5^\circ$} & \multicolumn{1}{c}{$10^\circ$} & \multicolumn{1}{c}{$20^\circ$}
      & \multicolumn{1}{c}{$0.03$} & \multicolumn{1}{c}{$0.06$} & \multicolumn{1}{c}{$0.1$} \\
    \midrule
    \multirow{2}{*}{Day}
      & Closer  & -20.34 & -19.33 & -17.10 & -13.12 & -18.32 & -15.09 & -8.41 \\
      & Farther & +14.32 & +12.49 & +9.63  & +4.19  & +11.29 & +5.39  & +1.16 \\
    \midrule
    \multirow{2}{*}{Night}
      & Closer  & -29.12 & -25.34 & -20.72 & -11.62 & -22.97 & -19.25 & -13.84 \\
      & Farther & +8.52  & +6.47  & +1.79  & -4.41  & +5.13  & +0.27  & -3.33 \\
    \bottomrule
  \end{tabular}
\end{table}

\subsection{\revision{}{Cross-Model and Cross-Task Transferability}}
\label{sec:appendix_transfer}

\revision{}{%
Using front camera sequences from nighttime nuScenes scenes, we evaluate whether scratch patterns optimized for one model continue to produce a depth shift on other models.
For each source model, we optimize the same scratch pattern throughout each target sequence, then apply these patterns to all six target models without re-optimization.
The evaluation covers three monocular depth estimation models, MonoDepth2, md4all, and DCPI-Depth, and three monocular 3D detectors, FCOS3D, PGD, and OVM3D.%
}

\revision{}{%
Table~\ref{tab:transfer_matrix} reports nighttime transfer results for the source-optimized scratch patterns.
For monocular depth estimation, each cell reports the transferred $\Delta\mathrm{RE}$.
For 3D detection, each cell reports $\Delta\mathrm{RE}$ together with the target loss rate.
The italicized diagonal values are same-model results recomputed within the same transfer evaluation; slight differences from Table~\ref{tab:monodepth} arise from stochastic optimization.%
}

\begin{table}[t]
  \centering
  \caption{\revision{}{%
% \by{do not use 'identify', just use 'be'}
% \ql{RESOLVED}
Cross-model and cross-task transfer at night. For each source-target pair, we report the larger-magnitude $\Delta\mathrm{RE}$ obtained from source scratches optimized for the closer and farther objectives. Monocular depth estimation cells report $\Delta\mathrm{RE}$, while 3D detection cells report $\Delta\mathrm{RE}$/TLR. All values are percentages. Rows are source models, and columns are target models. MD2, MD4, and DCPI are MonoDepth2, md4all, and DCPI-Depth. Italicized diagonal values are same-model reference results.%
}}
  \label{tab:transfer_matrix}
  \scriptsize
  \setlength{\tabcolsep}{3pt}
  \resizebox{\columnwidth}{!}{%
\revision{}{%
  \begin{tabular}{l *{3}{S[table-format=+2.1]} ccc}
    \toprule
    \multirow{2}{*}{} & \multicolumn{3}{c}{Monocular Depth Estimation} & \multicolumn{3}{c}{3D Detection} \\
    \cmidrule(lr){2-4}\cmidrule(lr){5-7}
      & \multicolumn{1}{c}{MD2} & \multicolumn{1}{c}{MD4} & \multicolumn{1}{c}{DCPI} & FCOS3D & PGD & OVM3D \\
    \midrule
    MD2    & \multicolumn{1}{c}{\textit{\tablenum[table-format=+2.1]{-28.9}}} & -16.3 & +10.6 & \tablenum[table-format=+1.1]{+1.0}/\tablenum[table-format=2.1]{31.7} & \tablenum[table-format=+1.1]{+0.7}/\tablenum[table-format=2.1]{24.3} & \tablenum[table-format=+1.1]{+0.9}/\tablenum[table-format=2.1]{14.5} \\
    MD4    & -19.1 & \multicolumn{1}{c}{\textit{\tablenum[table-format=+2.1]{-31.8}}} & +12.7 & \tablenum[table-format=+1.1]{+0.9}/\tablenum[table-format=2.1]{12.7} & \tablenum[table-format=+1.1]{+0.4}/\tablenum[table-format=2.1]{25.7} & \tablenum[table-format=+1.1]{+1.3}/\tablenum[table-format=2.1]{14.5} \\
    DCPI   & -16.0 & -23.7 & \multicolumn{1}{c}{\textit{\tablenum[table-format=+2.1]{+35.6}}} & \tablenum[table-format=+1.1]{+0.4}/\tablenum[table-format=2.1]{52.4} & \tablenum[table-format=+1.1]{-1.1}/\tablenum[table-format=2.1]{14.3} & \tablenum[table-format=+1.1]{-1.4}/\tablenum[table-format=2.1]{16.1} \\
    FCOS3D & -6.2 & +4.7 & +15.6 & \textit{\tablenum[table-format=+1.1]{+3.8}/\tablenum[table-format=2.1]{0.0}} & \tablenum[table-format=+1.1]{+1.9}/\tablenum[table-format=2.1]{20.8} & \tablenum[table-format=+1.1]{-1.8}/\tablenum[table-format=2.1]{10.9} \\
    PGD    & -11.0 & -0.4 & +13.0 & \tablenum[table-format=+1.1]{+1.0}/\tablenum[table-format=2.1]{17.0} & \textit{\tablenum[table-format=+1.1]{+3.6}/\tablenum[table-format=2.1]{1.4}} & \tablenum[table-format=+1.1]{-2.0}/\tablenum[table-format=2.1]{11.3} \\
    OVM3D  & -5.4 & +7.4 & +8.4 & \tablenum[table-format=+1.1]{+0.8}/\tablenum[table-format=2.1]{13.0} & \tablenum[table-format=+1.1]{-0.9}/\tablenum[table-format=2.1]{7.5} & \textit{\tablenum[table-format=+1.1]{-6.5}/\tablenum[table-format=2.1]{0.0}} \\
    \bottomrule
  \end{tabular}%
}
  }
\end{table}

\revision{}{%
\noindent\textbf{Transfer across depth models.}
Optimized scratch patterns continue to produce measurable depth shifts when applied to a different depth model.
For example, md4all$\rightarrow$MonoDepth2 reaches $-19.1\%$, while DCPI-Depth$\rightarrow$md4all reaches $-23.7\%$.
MonoDepth2 and md4all share a backbone and are primarily affected in the closer direction, whereas DCPI-Depth uses a different architecture and is primarily affected in the farther direction.
Under cross-model transfer, scratches applied to DCPI-Depth push its predictions farther, while scratches applied to MonoDepth2 and md4all pull their predictions closer.
Thus, measurable transfer occurs across models with different architectures and depth responses.%
}

\revision{}{%
\noindent\textbf{Transfer across tasks.}
Optimized scratch patterns also transfer between monocular depth estimation and 3D detection.
When scratches optimized on depth models are applied to detectors, the target loss rate reaches $13\%$ to $52\%$, even when $\Delta\mathrm{RE}$ among the detected targets remains close to zero.
In the opposite direction, detector-optimized scratches produce measurable depth shifts: FCOS3D$\rightarrow$DCPI-Depth reaches $+15.6\%$, while OVM3D$\rightarrow$md4all reaches $+7.4\%$.
Together, these results show that optimized scratch patterns retain measurable effects across model architectures and perception tasks.%
}

\subsection{Physical Experiment Setup}
\label{sec:appendix_physical_setup}

This subsection provides the camera, cover, and engraving details used in the physical evaluation.

\noindent\textbf{Camera setup.}
We mount two Basler acA1920-48gc cameras in a vertically aligned, stacked configuration on a tripod at approximately 140\,cm height.
\revision{}{%
The cameras' optical centers are separated by approximately 3\,cm.%
}
Each camera uses an Edmund Optics 8.5\,mm/F2.8 lens and a custom-cut 1\,mm acrylic cover.
One cover carries the attack scratch pattern, while the other remains clean and serves as the control.
We engrave the acrylic cover by hand with a utility knife.
The cameras capture synchronously, yielding paired clean and attacked views of the same scene.

\noindent\textbf{Why the cover rather than the lens.}
The removable cover provides a low-cost and replaceable surface for reproducing the attack without permanently damaging the lens.
It represents an accessible front optical surface, such as a protective cover, filter, or outer lens surface, through which light passes before reaching the camera sensor.

\begin{figure}[t]
  \centering
  \includegraphics[width=0.384\columnwidth]{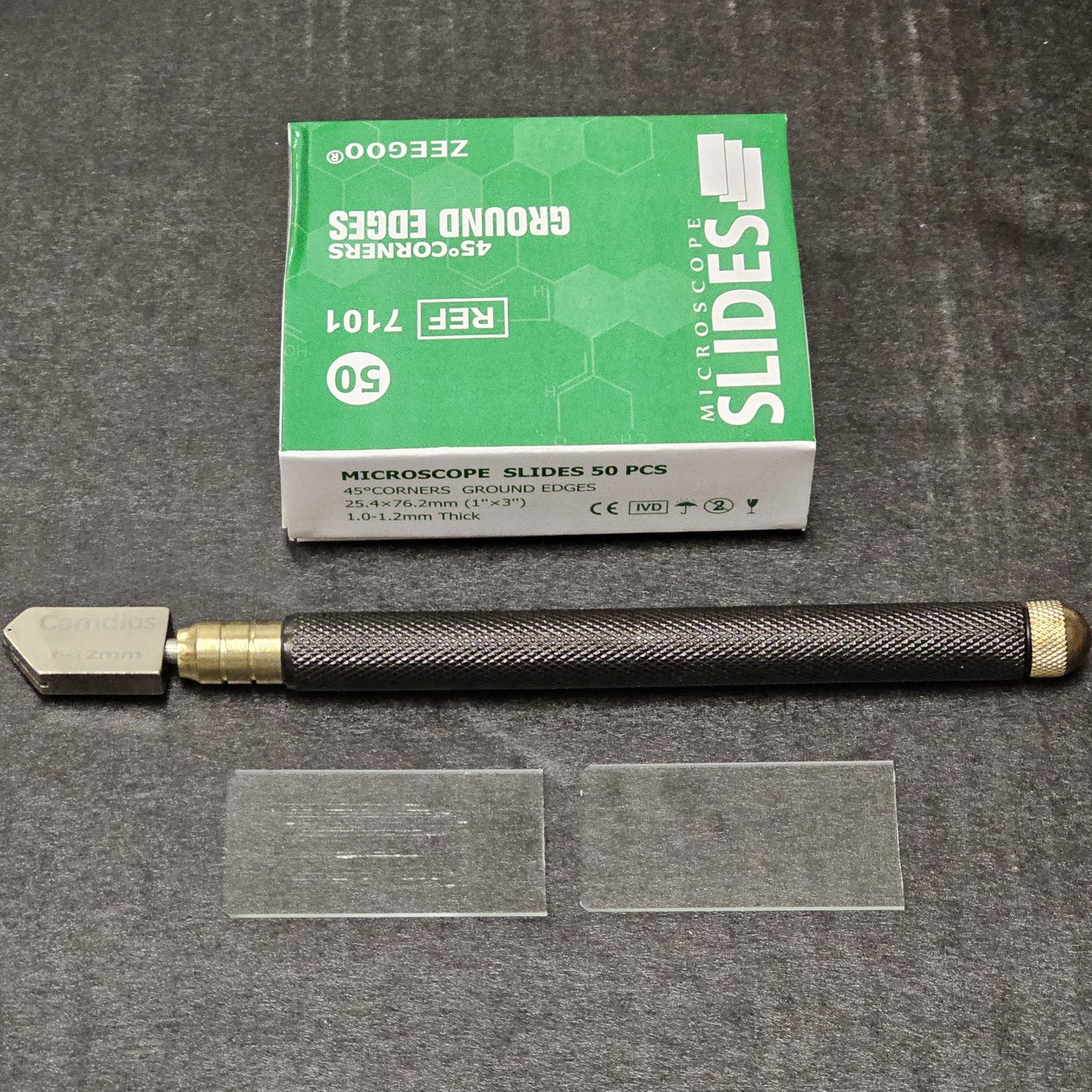}%
  \includegraphics[width=0.384\columnwidth]{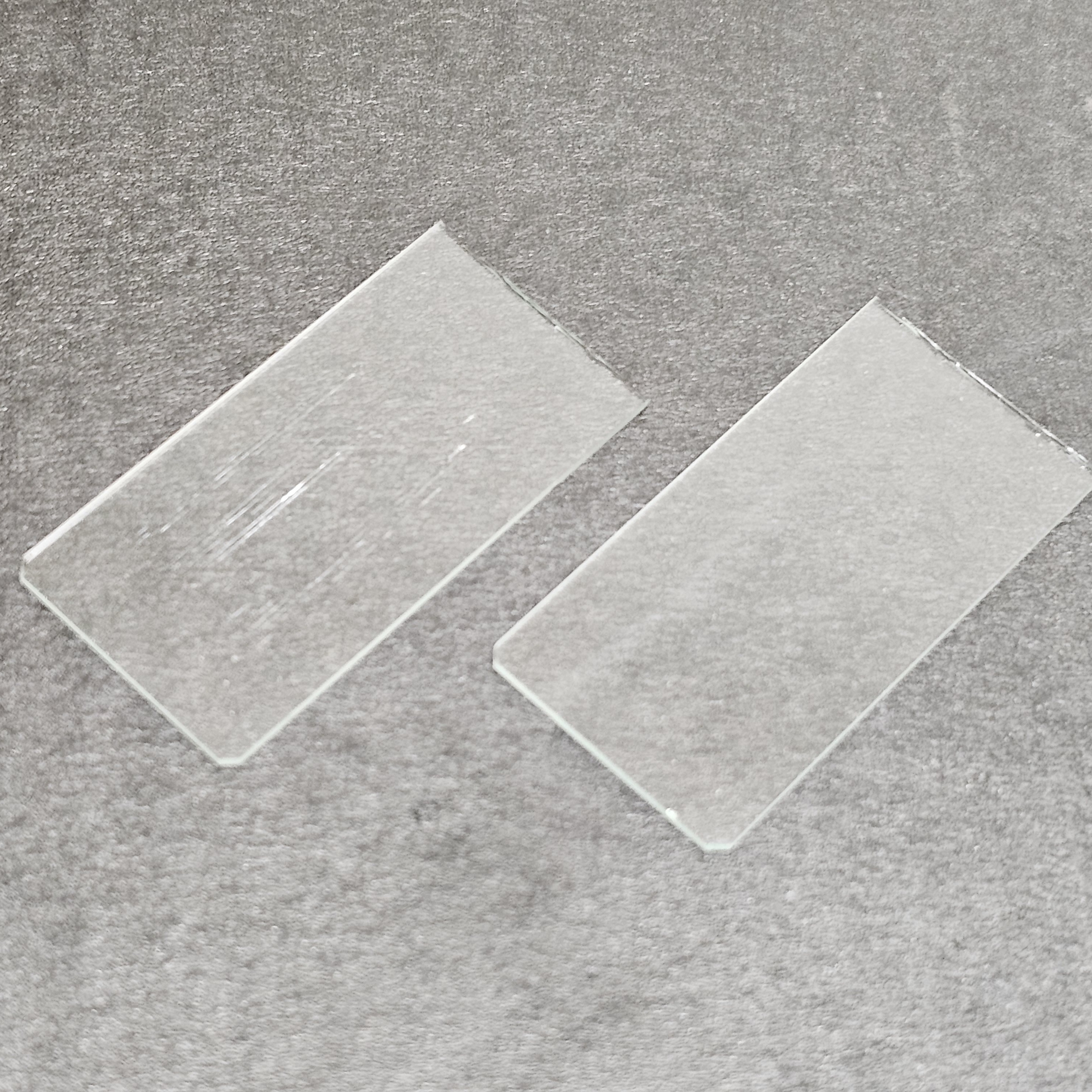}
  \caption{\revision{}{%
Glass-cover fabrication. The left photo shows two glass covers and a tungsten carbide rolling-wheel cutter; the engraved cover is on the left. The right photo shows a close-up of the engraved scratch pattern.%
}}
  \label{fig:glass_fabrication}
\end{figure}

\revision{}{%
\noindent\textbf{Engraving the scratch on glass.}
The glass cover is a standard biological microscope slide approximately 1\,mm thick, matching the thickness of the acrylic cover.
We engrave the scratch pattern by hand using a commercially available tungsten carbide rolling-wheel glass cutter.
Figure~\ref{fig:glass_fabrication} shows the glass covers and the engraving tool.%
}

\subsection{\revision{}{Attack on Stereo Depth Estimation}}
\label{sec:appendix_stereo}

\begin{figure}[t]
  \centering
  \includegraphics[width=\columnwidth]{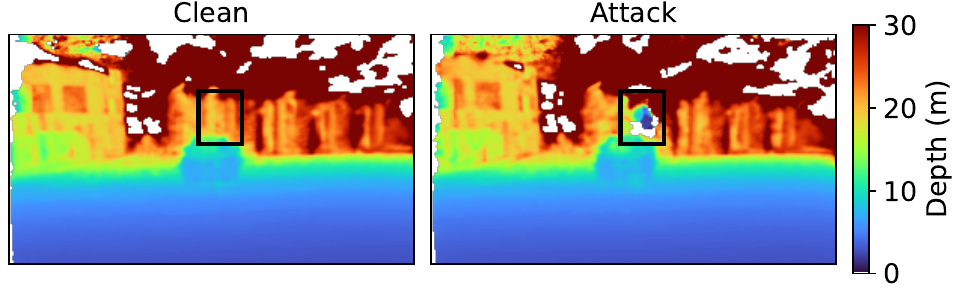}
  \caption{\revision{}{%
Effect of a scratched camera input on stereo depth. One camera has a scratched glass cover while the other remains clean. The panels show the average stereo depth under the clean and attack conditions. The marked region is the area affected by the attack.%
}}
  \label{fig:stereo_contamination}
\end{figure}

\revision{}{%
We evaluate whether a second camera view prevents the attack with a commercial stereo camera whose two integrated cameras jointly produce a depth estimate.

We use an Intel RealSense stereo camera to record the scene with clean covers on both cameras, then place a scratched glass cover over one camera while the other remains clean.
The camera and scene remain unchanged between the two recordings, and we compare the average predicted depth in the same image region.

Figure~\ref{fig:stereo_contamination} shows the outcome.
With clean covers, the stereo system predicts the marked region at a stable depth close to its measured distance.
After the scratched cover is placed over one camera, the same region becomes a dense false surface substantially closer to the camera.
Thus, contaminating one input can corrupt the fused stereo depth estimate despite the clean second view.%
}

\subsection{\proposed Attack against Countermeasures}
\label{sec:appendix_countermeasures}

\revision{}{%
This subsection evaluates flare removal, common image enhancement, and adversarial training against \proposed using md4all on the nuScenes dataset.%
}

\ifrevisionshowdiff
{
\color{revdelcolor}
\begin{table}[t]
  \centering
  \caption{Effectiveness of flare removal as a preprocessing defense. We evaluate md4all on front-camera target sequences. Columns report relative error (RE) on the clean input, attacked input, and attacked input after Flare7K++~\cite{dai2022_flare7k} or MFDNet~\cite{jiang2024_mfdnet}. All values are percentages.}
  \label{tab:defense_flare_removal}
  \small
  \setlength{\tabcolsep}{4pt}
  \begin{tabular}{l *{4}{S[table-format=+2.1]}}
    \toprule
    Condition
    & \multicolumn{1}{c}{Clean}
    & \multicolumn{1}{c}{Attack}
    & \multicolumn{1}{c}{Flare7K++}
    & \multicolumn{1}{c}{MFDNet} \\
    \midrule
    Day / Farther   & +6.6  & +18.1 & +11.7 & +10.1 \\
    Day / Closer    & +6.6  & -9.2  & -0.5  & +0.2  \\
    Night / Farther & +13.9 & +36.9 & +35.4 & +33.5 \\
    Night / Closer  & +13.9 & -18.5 & -15.9 & -13.0 \\
    \bottomrule
  \end{tabular}
\end{table}
}
\fi

% Keep the revised Clean RE + Delta RE layout: the original layout combines absolute attacked RE and attack-induced Delta RE under the same Delta RE label, both layouts use the same underlying results.
\begin{table}[t]
  \centering
  \caption{\revision{%
Adversarial training evaluation (MD4All, front camera, scene-specific optimization). Adv.\ Trained $\Delta\mathrm{RE}$ reports the adaptive-attack result against the hardened model.%
}{%
Effectiveness of adversarial training against an adaptive attack. We evaluate md4all and its fine-tuned counterpart on front-camera target sequences. Each model is reported as its unattacked bias (Clean RE) and the depth shift added by an attack optimized against that model ($\Delta\mathrm{RE}$). All values are percentages.%
}}
  \label{tab:adv_training}
  \small
  \setlength{\tabcolsep}{5pt}
\revision{%
  \begin{tabular}{lccc}
    \toprule
    Condition
    & \makecell{Clean\\$\Delta$RE (\%)}
    & \makecell{Attack\\$\Delta$RE (\%)}
    & \makecell{Adv.\ Trained\\$\Delta$RE (\%)} \\
    \midrule
    Day / Farther   & $+6.6$  & $+18.1$  & $+11.4$  \\
    Day / Closer    & $+6.6$  & $-9.1$   & $-13.5$  \\
    Night / Farther & $+13.9$ & $+36.9$  & $+20.2$  \\
    Night / Closer  & $+13.9$ & $-18.5$  & $-33.5$  \\
    \bottomrule
  \end{tabular}%
}{%
  \begin{tabular}{cc S[table-format=+2.1] S[table-format=+2.1] S[table-format=+2.1]}
    \toprule
    \multirow{2}{*}{Scene}
    & \multirow{2}{*}{Model}
    & \multicolumn{1}{c}{\multirow{2}{*}{Clean RE}}
    & \multicolumn{2}{c}{$\Delta$RE} \\
    \cmidrule(lr){4-5}
    & & & \multicolumn{1}{c}{Closer} & \multicolumn{1}{c}{Farther} \\
    \midrule
    \multirow{2}{*}{Day}
      & md4all     & +6.6 & -15.7 & +11.5 \\
      & Fine-tuned & +4.5 & -13.5 & +11.4 \\
    \midrule
    \multirow{2}{*}{Night}
      & md4all     & +13.9 & -32.4 & +23.0 \\
      & Fine-tuned & +19.8 & -33.5 & +20.2 \\
    \bottomrule
  \end{tabular}%
}
\end{table}

\noindent\textbf{Flare removal.}
Before md4all inference, Flare7K++~\cite{dai2022_flare7k} and MFDNet~\cite{jiang2024_mfdnet} preprocessing reduces the nighttime farther shift from $+23.0\%$ to $+21.5\%$ and $+19.6\%$, respectively.
\revision{}{%
In contrast, during the day, the two methods leave roughly $30\%-45\%$ of the depth shift across both attack directions.%
}

\revision{%
Figure~\ref{fig:flare_removal_qual} shows that the two methods alter the image differently.%
}{}
Flare7K++ leaves residual streak structure, while MFDNet suppresses legitimate light sources, including the vehicle headlights and a nearby lamp.
This removal also makes the target vehicle less distinct in the RGB image and changes the resulting depth prediction.

We benchmark both on an L40S GPU using $1600\times900$ images.
The md4all model requires $14.6\,\mathrm{ms}$ per frame.
Adding MFDNet increases end-to-end latency to approximately $50\,\mathrm{ms}$ per frame, while Flare7K++ increases it to approximately $170\,\mathrm{ms}$ per frame.
\revision{%
Thus, both defenses retain a substantial attack effect and add considerable processing cost.%
}{%
Thus, both defenses reduce but do not eliminate the measured depth shifts and add considerable processing cost.%
}

\ifrevisionshowdiff
\begin{figure}[t]
  \centering
  \includegraphics[width=\linewidth]{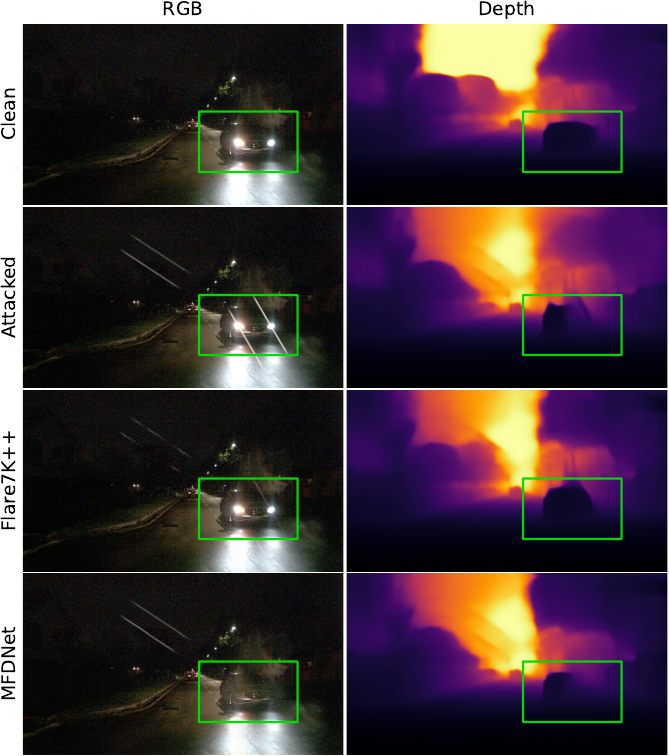}
  \caption{{\color{revdelcolor}Qualitative effects of flare removal on an attacked input.
  Rows show the clean input, attacked input, Flare7K++~\cite{dai2022_flare7k} output, and MFDNet~\cite{jiang2024_mfdnet} output.
  Columns show the RGB image with the target box and the corresponding md4all depth map for the nighttime farther-direction setting.}}
  \label{fig:flare_removal_qual}
\end{figure}
\fi

\revision{}{%
\noindent\textbf{Common image enhancement.}
We apply pretrained Zero-DCE~\cite{guo2020_zero_dce} and Retinexformer~\cite{cai2023_retinexformer} for low-light enhancement, NAFNet~\cite{chen2022_nafnet} for denoising, and CLAHE~\cite{zuiderveld1994_clahe} for contrast enhancement symmetrically to clean and attacked md4all front-camera inputs.
We define the clean depth shift as $\mathrm{RE}^{\mathrm{enh}}_{\mathrm{clean}}-\mathrm{RE}_{\mathrm{clean}}$ and the remaining attack effect as $\mathrm{RE}^{\mathrm{enh}}_{\mathrm{attack}}-\mathrm{RE}^{\mathrm{enh}}_{\mathrm{clean}}$.
At night, the remaining farther effect is $+22.6\%$ with Zero-DCE, $+15.9\%$ with Retinexformer, $+22.2\%$ with NAFNet, and $+16.9\%$ with CLAHE, compared with $+23.0\%$ without enhancement.
Retinexformer also produces a $+7.8\%$ clean depth shift, while CLAHE increases the remaining nighttime closer effect from $-32.4\%$ to $-37.8\%$.
Thus, none of the evaluated enhancement methods neutralizes the attack.%
}

\revision{%
\noindent\textbf{Adversarial training: detailed analysis.}
We fine-tune MD4All on the nuScenes training split with online scratch augmentation ($p{=}0.5$, geometry-based streak renderer with uniformly sampled parameters, 10 epochs), then evaluate under an adaptive attack: the same per-sequence SLASH protocol (CMA-ES, two scratches, front camera) re-optimised against the defended model with full knowledge of the defense.

Fine-tuning increases the clean $\Delta\mathrm{RE}$ from $+13.9\%$ to $+19.8\%$ in the nighttime condition, confirming the known clean-accuracy cost.
Against the adaptive attacker, gains are marginal: for night/farther, the attack-induced effect decreases from $+23.0\,\mathrm{pp}$ to $+20.2\,\mathrm{pp}$; for night/closer it worsens slightly ($-32.4\,\mathrm{pp} \to -33.5\,\mathrm{pp}$); daytime conditions are nearly unchanged.
The adaptive optimizer is thus able to fully circumvent the defense, consistent with the known limitation that adversarial training provides robustness only against the artifact distribution seen during training.
A scratch optimized by SLASH to lie outside that distribution would likely remain effective.%
}{%
\noindent\textbf{Adversarial training.}
We fine-tune md4all on the nuScenes training split for 10 epochs using online scratch augmentation with probability $p=0.5$ and uniformly sampled renderer parameters.
We then evaluate an adaptive attacker that re-optimizes the scratch pattern against the fine-tuned model under the same scene-specific optimization setting.

Table~\ref{tab:adv_training} reports clean RE and attack-induced $\Delta\mathrm{RE}$ for the original and fine-tuned models.
Fine-tuning improves daytime clean RE from $+6.6\%$ to $+4.5\%$ and increases nighttime clean RE from $+13.9\%$ to $+19.8\%$.
Under the adaptive attack, nighttime $\Delta\mathrm{RE}$ reaches $+20.2\%$ in the farther direction and $-33.5\%$ in the closer direction, compared with $+23.0\%$ and $-32.4\%$ for the original model.
Thus, scratch augmentation does not neutralize an attack re-optimized against the fine-tuned model.%
}

\end{document}